\documentclass[conference]{IEEEtran}
\usepackage[english]{babel}

\usepackage{glossaries}
\newglossaryentry{ctrnn}{
    name={CtRNN},
    description={Continuous Time Recurrent Neural Network},
}

\RequirePackage[l2tabu, orthodox]{nag}

\usepackage{pdflscape}

\usepackage{newtxtext,newtxmath}

\usepackage{textcomp}
\usepackage{graphics}

\usepackage{comment}
\usepackage{cite}

\usepackage{booktabs}
\newcommand{\ra}[1]{\renewcommand{\arraystretch}{#1}}

\usepackage{multirow}

\usepackage{caption}
\usepackage{comment}

\usepackage{tabularx}
\usepackage{xcolor}
\usepackage{needspace}
\usepackage[pdftex]{graphicx}
\graphicspath{{./figures/}{./resources/}}
\DeclareGraphicsExtensions{.pdf,.png,.jpg,.jpeg}
\usepackage{wrapfig}
\usepackage{tikz}
\usetikzlibrary{shapes,arrows}
\usepackage{pgfplots}
\pgfplotsset{compat=1.9}

\usepackage{amsmath}
\usepackage{mathtools}

\ifCLASSOPTIONcompsoc
  \usepackage[caption=false,font=normalsize,labelfont=sf,textfont=sf]{subfig}
\else
  \usepackage[caption=false,font=footnotesize]{subfig}
\fi

\usepackage{url}

\usepackage{xspace}
\makeatletter
\DeclareRobustCommand\onedot{\futurelet\@let@token\@onedot}
\def\@onedot{\ifx\@let@token.\else.\null\fi\xspace}

\makeatother

\usepackage{tcolorbox}

\makeatletter

\def\ps@IEEEtitlepagestyle{%
     \def\@oddfoot{\mycopyrightnotice}%
     \def\@evenfoot{}%
}
\def\mycopyrightnotice{%
     {\begin{minipage}{2\linewidth}\footnotesize\bfseries \copyright
2024 IEEE.  Personal use of this material is permitted.  Permission from IEEE must be obtained for all other uses, in any current or future media, including reprinting/republishing this material for advertising or promotional purposes, creating new collective works, for resale or redistribution to servers or lists, or reuse of any copyrighted component of this work in other works.\hfill\end{minipage}}
     \gdef\mycopyrightnotice{}
}
\begin{document}

\title{Energy Efficient Deep Multi-Label ON/OFF Classification of  Low Frequency Metered Home Appliances}

\author{\IEEEauthorblockN{%
    An\v{z}e~Pirnat,  
    Bla\v{z}~Bertalani\v{c},  
    Gregor~Cerar,  
    Mihael~Mohor\v{c}i\v{c}, 
    and Carolina~Fortuna%\IEEEauthorrefmark{1}
}\IEEEauthorblockA{%
    Department of Communication Systems, Jo\v{z}ef Stefan Institute, Slovenia.\\
    %\IEEEauthorrefmark{2}Faculty of Electrical Engineering, University of Ljubljana, Slovenia.
}%
%Corresponding~author: ???
ap6928@student.uni-lj.si, \{blaz.bertalanic, gregor.cerar, mihael.mohorcic, carolina.fortuna\}@ijs.si
}




\maketitle

\begin{abstract}
%% What is important

%being able to know which devices are on. good for energy efficiency -> ECO, can help in assisted living . . .

%Non-intrusive load monitoring (NILM) is the process of obtaining appliance-level data from users' total electricity consumption data. These data can be of great benefit, especially in demand response applications.

%Non-Intrusive Load Monitoring consists in estimating the power consumption or the states of the appliances using electrical parameters acquired from a single metering point.

%Commercial load is an essential demand-side resource. Monitoring commercial loads helps not only commercial customers understand their energy usage to improve energy efficiency but also helps electric utilities develop demand-side management strategies to ensure stable operation of the power system

Non-intrusive load monitoring (NILM) is the process of obtaining appliance-level data from a single metering point, measuring total electricity consumption of a household or a business. Appliance-level data can be directly used for demand response applications and energy management systems as well as for awareness raising and motivation for improvements in energy efficiency. 
%Beside that it can also help electric companies develop demand-side management strategies to ensure stable operation of the power system.
%Popular approaches for NILM classification are classical machine learning (ML) and deep learning (DL) algorithms which perform very well.
Recently, classical machine learning and deep learning (DL) techniques became very popular and proved as highly effective for NILM classification, but with 
%% What is missing/not properly solved?
%Tests aren't realistic and often show results that are missleading in the context of industry but sensible in the context of research where they just try to prove their approach is better. Energy efficiency of the approach isn't always considered.
%However, in many cases testing of classification algorithms is done in a way that is sensible in the context of research, thus good for comparing new approaches, but gives results that can be missleading in the context of real world use. 
%However, with the increasing complexity these methods become computationally very intensive and energy hungry, both for their training and subsequent operation. Which is a major downside when we are trying to lessen the carbon footpritn. Additionally, current performance testing methodology employed can yield misleading results when applied to real-world scenarios.
the growing complexity these methods are faced with significant computational and energy demands during both their training and operation. 
%, posing a challenge to reducing carbon footprints. Moreover, the existing performance testing methodology may yield unrealistic results.
%% How do we contribute to solving it
%In this paper we propose our DL model CtRNN with only 70.2 percentage points of the carbon footprint of the state of the art, to contribute to energy efficiency.
%On REFIT, CtRNN is 11.03 percentage points better on average on test A and 11.32 percentage points on test B, compared to state of the art DL model VGG11. And on UK-DALE its 9.4 percentage points better on average on test A and 8.07 percentage points on test B. 
In this paper, we 
% 1st contribution
introduce a novel DL model aimed at enhanced multi-label classification of NILM with improved computation and energy efficiency. 
% 2nd contribution
We also propose an evaluation methodology for comparison of different models using data synthesized from the measurement datasets so as to better represent real-world scenarios.
% Result of 1st&2nd contribution
Compared to the state-of-the-art, the proposed model has its energy consumption reduced by more than 23\% while providing on average approximately 8 percentage points in performance improvement when evaluating on data derived from REFIT and UK-DALE datasets. 
% Result of 3rd contribution
We also show a 12 percentage point performance advantage of the proposed DL based model over a random forest model and observe performance degradation with the increase of the number of devices in the household, namely with each additional 5 devices, the average performance  degrades by approximately 7 percentage points.
\end{abstract}


\maketitle

\section{Introduction}
\label{sec:intro}

% problem: climate change
Climate change represents a formidable challenge, and mitigating its impacts requires a concerted effort to maintain the increase in global average temperature below 1.5\,$^{\circ}$C relative to pre-industrial levels.
% solution 1: cutting down on electricity production
Electrical energy production is estimated to contribute more than 40\,\% of the total CO$_2$ equivalent produced by humankind\footnote{%https://planete-energies.com/en/medias/close/electricity-generation-and-related-co2-emissions
https://tinyurl.com/electricity-production-CO2-1 (accessed 4.3.2024)}\textsuperscript{,}\footnote{%https://world-nuclear.org/information-library/energy-and-the-environment/carbon-dioxide-emissions-from-electricity.aspx
https://tinyurl.com/electricity-production-CO2-2 (accessed 4.3.2024)}. So, some of the necessary steps in mitigating climate change are to reduce energy consumption and subsequently its production as well as to increase the share of renewable energy sources\footnote{https://tinyurl.com/renewable-energy-doubled (accessed 4.3.2024)} that produce far less CO$_2$ equivalent compared to traditional power plants that burn fossil fuels.
% solution 2: energy with less CO2
However, the renewable energy sources are mostly dependant on external conditions such as wind, sun, etc. and are thus less predictable and pose a challenge to the stability and reliability of the electrical power system~\cite{Ali2020}. To solve this problem we have to work with the concept of demand response; change the electrical power consumption to better match the demand with the supply~\cite{Aghaei2013}. Because of demand response, efforts are being made to monitor and manage energy consumption more effectively in residential building, which makes monitoring the activity of devices (ON/OFF events) relevant~\cite{Gopinath2020}.

% ||||||||||||||||||||||||||||||||||||||||||||| SECOND PARA ||||||||||||||||||||||||||||||||||||||||||||||||||||||||||||

% NILM useful for solution 2
Monitoring each device separately is costly and invasive since it requires an installation of an electricity meter on each appliance. As an alternative, a non-intrusive load monitoring (NILM) supported with disaggregation methods is able to reach the same result with just one electricity meter per household and is thus much more economical~\cite{Hart1992_OG_NILM_paper}. NILM is the process of obtaining appliance-level data from a single metering point measuring total electricity consumption of a household or a business. By subsequent processing, it is possible to decompose NILM data into individual components, and by classification we can determine the state (ON/OFF) of devices and thus monitor their activity for demand response applications. 
% NILM useful for solution 1
In Europe, households consume 27.4\,\% of all electricity produced\footnote{%https://tinyurl.com/household-consumption-statstic
https://tinyurl.com/home-consumption-statistic (accessed 4.3.2024)}.
%~\footnote{tinyurl.com/eurostat-statistic}.
Thus cutting down on their consumption would play an important role in relieving our carbon footprint. As several research studies have shown, if given real-time feedback on their electricity savings residents achieve a more comprehensive understanding of their electrical consumption and develop more energy aware behaviour. Consequently they consume 12\,\% less electricity than they would normally~\cite{ehrhardt2010advanced}. With classification on NILM we can provide feedback on the device activity and help that way.

% ||||||||||||||||||||||||||||||||||||||||||||| THIRD PARA ||||||||||||||||||||||||||||||||||||||||||||||||||||||||||||

\begin{figure*}[ht!]
    \centering
    \includegraphics[width=\linewidth]{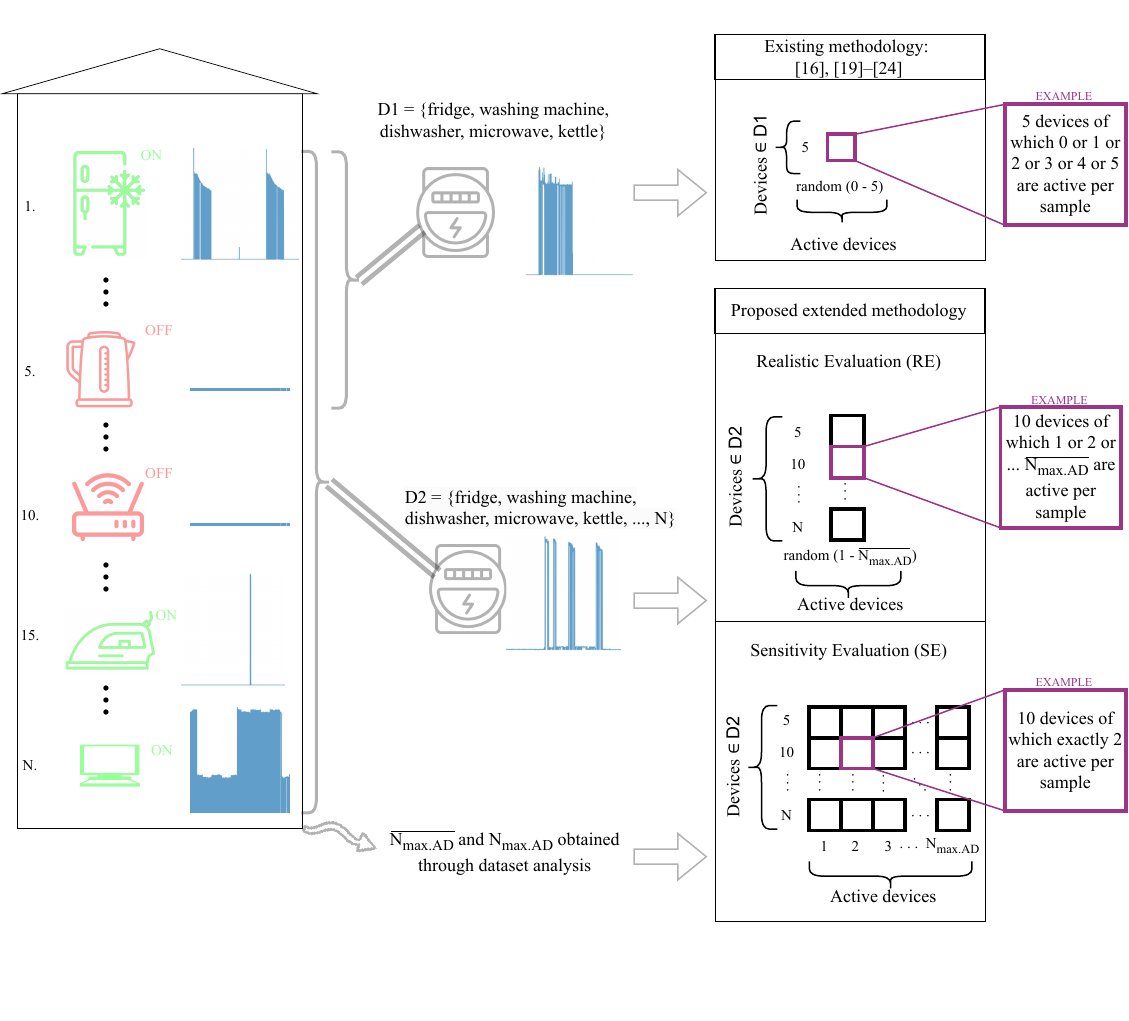}
    \caption{The proposed evaluation methodology of groups SE and RE of mixed datasets compared to the proposed evaluation methodology.}
    \label{fig:AB-Explanation}
\end{figure*}
In the described application areas for classification on NILM there can be more than one device active at a time. Thus, the best approach to determine activity states of the appliances is multi-label classification, where the state of the appliance is used as the class label and the recorded readings from a single meter on the household as input samples. Multi-label classification has been attempted on NILM with numerous methods that can be divided into two categories. The first category includes single channel source separation techniques such as matrix factorization~\cite{rahimpour2017non}, sparse coding~\cite{kolter2010energy}, dictionary learning~\cite{singh2017deep}, and non-negative tensor factorization~\cite{figueiredo2013electrical}, while the second category comprises machine learning approaches such as support vector machines (SVM)~\cite{chui2018energy}, random forests (RF)~\cite{Wu2019}, decision trees (DT)~\cite{Buddhahai2018} and deep neural networks~\cite{Zhou2022, Massidda2020, Bertalanič2022, Tanoni2022}.

\subsection{Contributions and Paper Organization}
\label{sec:contributions}

% ||||||||||||||||||||||||||||||||||||||||||||| FOURTH PARA ||||||||||||||||||||||||||||||||||||||||||||||||||||||||||||

% |||||||||||||||||||||||||||||||||||||||||| CONTRIBUTIONS ||||||||||||||||||||||||||||||||||||||||||||||||||||
In this paper we are concerned with energy efficient ON/OFF classification of NILM data aimed at decreasing the overall energy consumption, which includes the investigation whether more complex and accurate DL approaches outweigh simpler and less consuming classical ML approaches. 
% added motivation for energy efficiency
Improved energy consumption, encouraged for aforementioned ecological reasons, is caused by the reduction in computational cost which is highly encouraged by the cloud computing community~\cite{Asres2022} and seen as a necessity in the field~\cite{Angelis2022}.

We propose a new DL architecture that is inspired by the VGG family of architectures and RNNs for the multi-label device activity classification on NILM data. We prove that its performance is better than state of the art CNNs and CRNNs and state of the art classical ML algorithms, while being much more energy efficient compared to the  state of the art DL models. 

Classical ML and DL models such as used in \cite{Tanoni2022, Langevin2022, Pan2020, D'Incecco2020, Wang2022, Zhang2018} and \cite{Kelly2015} are %mostly %tested %in a way that is better suited for the comparison with other models and not in a way that would give realistic results of performance in real-world conditions. 
trained and evaluated with only five particular devices, which occur commonly in different datasets and houses within. This method is well suited for the comparison with other models, but lacks the ability to reflect realistic results of performance as it fully disregards all other devices. Those 5 devices are chosen because they each draw large proportion of energy and represent various different power signatures. As stated in \cite{kelly2015neural} that can be especially problematic since all devices with smaller energy consumption specifically, are thus disregarded, while modern homes tend to have a lot of them. Moreover, different problem types on NILM data such as disaggregation and appliance classification already employ much more than 5 devices \cite{raiker2020energy, Ciancetta2021, Chen2022, Zhou2022}. The methodology should thus be extended to more devices and it should be done according to specifics of the dataset to make it as close to real-world examples as possible.

To this end, this work examines a novel methodology depicted in Figure \ref{fig:AB-Explanation} that more accurately represents the performance of models in practical use cases. The extended methodology complements existing methodologies  \cite{Tanoni2022, Langevin2022, Pan2020, D'Incecco2020, Wang2022, Zhang2018, Kelly2015}, by creating a so-called Realistic evaluation dataset and an Sensitivity evaluation dataset. The existing methodologies typically assume that there are 5 devices per household and any of these can be active in the observation window. In the Realistic evaluation dataset we allow for households to have a varying number of active devices, say 5, 10, 15, ... N devices, of which any random combination can be active. In the Sensitivity evaluation dataset we further extend the possible combinations by allowing a household to have varying number of devices (i.e. 5, 10, 15, ... N) however we study what happens when only 1, only 2, etc. are active at a time separately - thus enabling the assessment of the sensitivity to various combinations of devices. 

% mention we compare on both methodologies.
We evaluate our model on both the established evaluation methodology and the proposed evaluation methodology to assure its performance.

Our main contributions are as follows:
\begin{itemize}
    \item
    We propose a novel \textit{Convolutional transpose Reccurrent Neural Network} (CtRNN)  architecture focusing on reduced computational complexity which offers superior performance compared to the existing state of the art architectures with an average improvement of approximately 8 percentage points on mixed datasets derived from REFIT and UK-DALE and more than 23\,\% lesser energy consumption, making it a more sustainable solution.
    \item 
    We propose a novel evaluation methodology that begins with a dataset analysis and involves generating two groups of mixed datasets that are utilized for both training and testing. By taking into account the unique properties of the original dataset when generating mixed datasets, our approach results in a more realistic evaluation of model performance, more closely reflecting real-world scenarios.
    \item
    We perform a comprehensive analysis taking into account performance and energy efficiency of compared approaches for NILM ON/OFF classification. We observe an average of approximately 7 percentage point drop in F1 score for each   5 newly added devices in the household.
\end{itemize}

This paper is organized as follows. Section~\ref{sec:related_work} analyzed related work while Section~\ref{sec:setup} provides the problem statement and elaborates   on methodological aspects. Section~\ref{sec:model} presents the proposed model and Section~\ref{sec:evaluation} provides a comprehensive evaluation with Section~\ref{sec:conclusion} concluding the paper.

\begin{table*}[htbp]
    \ra{1.3}
    \caption{Comparison of results from other related works.}
    \label{tab:RelatedWork}
    \centering
    \resizebox{\textwidth}{!}{
    \centering
    \begin{tabular}{l|c|c|c|c|c}
        \toprule
        \bfseries Work %& \bfseries Problem type 
        & \bfseries Problem Type & \bfseries Approach Type & \bfseries Approach  & \bfseries Datasets  & \bfseries Devices no. \\
        \midrule
        Tabatabaei~\textit{et al.}~\cite{Tabatabaei2017}
        & ON/OFF classification
        & Classic ML 
        & RAkEL, MLkNN 
        & REDD (LF)
        & up to 5
        \\
        Raiker~\textit{et al.}~\cite{raiker2020energy}
        & Disaggregation
        & Classic ML 
        & fHMM
        & DRED, AMPds, BLUED, UK-DALE, WHITED, PLAID (LF\&HF)
        & up to 11
        \\ 
        Wu~\textit{et al.}~\cite{Wu2019}
        & ON/OFF classification
        & Classic ML 
        & RF
        & BLUED (HF)
        & 5
        \\
        Singh~\textit{et al.}~\cite{Singh2022}
        & ON/OFF classification
        & Classic ML 
        & SRC
        & REDD, Pecan Street (LF)
        & 4
        \\\hline
        \c{C}imen~\textit{et al.}~\cite{Cimen2020}
        & Disaggregation
        & DL 
        & AAE
        & UK-DALE, REDD (LF)
        & 5
        \\
        Ciancetta~\textit{et al.}~\cite{Ciancetta2021}
        & Appliance class.
        & DL 
        & CNN
        & BLUED (HF)
        & 34
        \\
        Chen~\textit{et al.}~\cite{Chen2022}
        & Appliance class.
        & DL 
        & TSCNN
        & PLAID, WHITED (HF)
        & up to 15
        \\
        Zhou~\textit{et al.}~\cite{Zhou2022Journal}
        & Appliance class.
        & DL 
        & SNN
        & PLAID (HF)
        & 11
        \\
        Yin~\textit{et al.}~\cite{Yin2023}
        & Appliance class.
        & DL 
        & DCNN
        & custom (HF)
        & up to 5
        \\
        Tanoni~\textit{et al.}~\cite{Tanoni2022}
        & ON/OFF classification
        & DL 
        & CRNN
        & REFIT, UK-DALE (LF)
        & 5
        \\
        Langevin~\textit{et al.}~\cite{Langevin2022}
        & ON/OFF classification
        & DL 
        & CRNN
        & REFIT, UK-DALE (LF)
        & 5
        \\
        \midrule
        This work
        & ON/OFF classification
        & DL 
        & CtRNN
        & REFIT, UK-DALE (LF)
        & up to 54
        \\
    \bottomrule
    \end{tabular}
    }
\end{table*}

\section{Related Work}
\label{sec:related_work}
%% RELATED WORK
%Literature review (usually several paragraphs):
%    Summarize the relevant literature on your topic
%    Describe the current state of the art
%    Note any gaps in the literature that your study will address

% This should be improved
In this section, we present related work focusing on multi-label classification on NILM with the use of classical machine learning (ML) algorithms and deep learning (DL) techniques. % brief explanation of the table I
To provide a comprehensive overview of the state-of-the-art in this area, we have compiled Table~\ref{tab:RelatedWork} summarizing selected more important references to prior work, including the type of the problem addressed, the approach used, the number and name of datasets utilized, and the number of devices involved in each study, however in the following subsections when discussing some of these specific aspects we also refer to some further relevant works.

\subsection{NILM Problem Type}
\label{subsec:NilmProblemType}

The second column in Table~\ref{tab:RelatedWork} demonstrates that state-of-the-art approaches for NILM can be categorized into three distinct types: disaggregation, ON/OFF classification, and appliance classification. The disaggregation problem is focused on decomposing the NILM signal into individual components that correspond to distinct power signatures of active appliances~\cite{raiker2020energy, Cimen2020}. The ON/OFF classification of appliances aims to determine which devices are active and which inactive in an aggregated power signal %This problem assumes that the disaggregated signals are already available, and as such, it builds on the results of the first type of problem. 
\cite{Tabatabaei2017,Wu2019,Singh2022,Tanoni2022,Langevin2022}. The appliance classification problem also assumes that the disaggregated signals are accessible and intends to classify the devices that generated each unique power signature extracted from the NILM signal~\cite{Ciancetta2021, Chen2022, Zhou2022Journal, Yin2023}.

The focus of this paper is on the ON/OFF classification problem type, which pertains to the identification of the activity state of individual appliances from an aggregated power signal without requiring prior disaggregation.

\subsection{Methods for Solving NILM Problems}
\label{subsec:NILMMethods}

As the analysis of the related work shows, the approaches to solving NILM involve either a two stage process in which first disaggregation is done that is followed by classification that performs automatic appliance identification or both, or a one stage process in which ON/OFF classification is done directly on aggregated data. %It can also be seen from the second column of~\ref{tab:RelatedWork} that most research has focused on solving the ON/OFF classification problem. 
In the last few years, several classic ML and DL methods have been proposed in this area. 

Classic ML algorithms utilized in reviewed work are Random k-labELset (RAkEL)~\cite{Li2015, Tabatabaei2017, Wu2019}, factorial Hidden Markov Model (fHMM)~\cite{raiker2020energy}, Random Forrest (RF)~\cite{Wu2019, Rehmani2021}, Sparse Representation based Classification (SRC)~\cite{Singh2022}, Classification And Regression Tree (CART)~\cite{Rehmani2021}, Extra Tree (ET)~\cite{Rehmani2021}, k-Nearest Neighbors (kNN)~\cite{Rehmani2021, Wu2019}, Linear Discrimination Analysis (LDA)~\cite{Rehmani2021} and Na\"ive Bayes (NB)~\cite{Rehmani2021}.

The latest state-of-the-art approaches for NILM are based on DL algorithms, as shown in Table~\ref{tab:RelatedWork}. In the reviewed works, Convolutional Neural Networks (CNN)~\cite{Ciancetta2021,Chen2022,Yin2023} and Recurrent CNNs (CRNN)~\cite{Bertalanič2022, Tanoni2022, Langevin2022} are the most common choice. However, a variety of other algorithms are also used, e.g. Adverserial Auto Encoders (AAE)~\cite{Cimen2020}, a custom archtecture TTRNet~\cite{Zhou2022} and Spiking Neural Network (SNN)~\cite{Zhou2022Journal}.

Utilizing the fully supervised learning method, Wu~\textit{et al.}\cite{Wu2019} conducted an experiment to evaluate various classical machine learning algorithms for multi-label classification of NILM data for load identification. Their findings indicate that Random Forest (RF) outperforms other learning algorithms. Similarly, Rehmani \textit{et al.}~\cite{Rehmani2021} demonstrated that computationally intensive deep learning (DL) algorithms, such as CNN and RNN, were not required for their particular datasets, as already classical machine learning algorithms, such as kNN and RF, yielded accuracy of 99\,\%. However, openly available and well documented REFIT and UK-DALE datasets, recently used in many reference works as well as in this study, do not exhibit suitable performance with the classical machine learning and are used with DL models. % was deemed necessary due to their superior performance compared to the classical machine learning approach of RF. 
To address the high computational complexity and energy consumption associated with DL models, we designed a novel DL architecture based on the principles established in our previous work \cite{Pirnat2022}. Additionally, to ensure consistency with the latest research, we compared our approach to works by Langevin \textit{et al.} \cite{Langevin2022} and Tanoni \textit{et al.} \cite{Tanoni2022}, who also utilized the same datasets of REFIT and UK-DALE. Furthermore, we also considered the findings of Ahajjam \textit{et al.} \cite{Ahajjam2021}, who discovered that the optimal signal length varies across datasets, and hence, we adopted the same signal length as Tanoni \textit{et al.} \cite{Tanoni2022}.

\begin{figure*}[ht]
    \centering
    \includegraphics[width=\linewidth, clip, trim={0 0 0 0}]{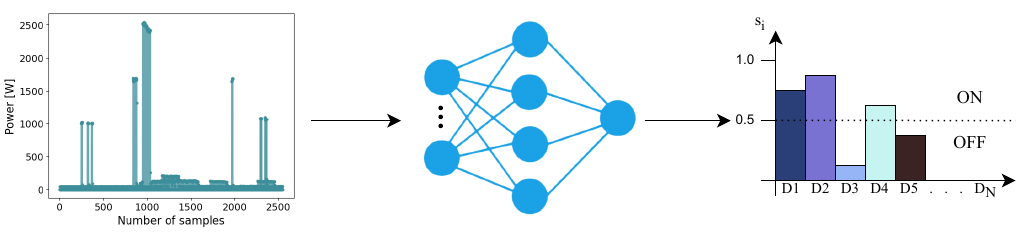}
    \caption{Classification of devices as active or inactive based on the household NILM data using classical ML or DL model to obtain $s_i$ for each device; devices with $s_i > 0.5$ are classified as active, the others are classified as inactive.}
    \label{fig:problem}
\end{figure*}

\subsection{NILM Data for ML Model Training}
\label{subsec:NILMdata}

As per the 5th column in Table~\ref{tab:RelatedWork}, the related works employed low-frequency (LF) and high-frequency (HF) datasets. %Nonetheless, it is noteworthy that LF datasets were more commonly employed. The prevalence of non-intrusive load monitoring (NILM) techniques on LF data in these studies may be attributed to the European Union and UK technical specifications for smart meters, which suggest a sampling rate of around 10 secondsfootnote{https://tinyurl.com/low-frequency-meters} for units installed in typical households. 
The European Union and UK technical specifications suggest the use of LF smart meters with a sampling rate of around 10 seconds\footnote{https://tinyurl.com/SMIP-E2E-SMETS2 (accessed 4.3.2024)} for units installed in typical households. To circumvent the need to purchase and install new HF smart meters, and instead utilize the existing LF meters whose readings are already available via the COSEM interface classes and OBIS Object Identification System\footnote{https://tinyurl.com/COSEM-interface-classes (accessed 4.3.2024)}, this paper proposes the development of an ON/OFF classification model for LF meters. %This model is directly compatible with the disaggregation methods described in lines 9 and 10 of the table in terms of sampling rate and functionality.

Typically, the number of devices employed in different works is fixed, with the exception of Raiker \textit{et al.} \cite{raiker2020energy}, Chen \textit{et al.} \cite{Chen2022} and Yin \textit{et al.} \cite{Yin2023}, who utilized up to 11, 15 and 5 devices, respectively. In this work, however, we utilize a flexible range of up to 54 devices.

\subsection{Energy Consumption}
\label{subsubsec:CO2}

The energy consumption of hardware used for running DL models and the resulting energy consumption has only recently become a growing concern in the community. In~\cite{Gigi2020}, Hsueh conducted an analysis of the energy consumption of ML algorithms and found that convolutional layers, operating in three dimensions, consume significantly more power compared to fully connected layers, which operate in two dimensions. Verhelst~\textit{et al.}~\cite{Verhelst2017} delved into the complexity of CNNs and explored hardware optimization techniques, particularly for the Internet of Things (IoT) and embedded devices. Another study by Garcia~\textit{et al.}~\cite{Garcia2019} surveyed the energy consumption of various models and proposed a taxonomy of power estimation models at both software and hardware levels. They also discussed existing approaches for estimating energy consumption, noting that using the number of weights alone is not accurate enough. They suggested that a more precise calculation of energy consumption requires the calculation of either FLOPs or multiply-accumulate operations.

%\begin{figure*}[htb]
%    \centering
%\includegraphics[width=\linewidth, clip, trim={0 0 0 10}]{figs/problemGrey.pdf}
%    \caption{Classification of devices as active or inactive based on the household NILM data using classical %ML or DL model to obtain $s_i$ for each device; devices with $s_i > 0.5$ are classified as active, the others %are classified as inactive.}
%    \label{fig:problem}
%\end{figure*}

\section{Problem Statement and Methodology}
\label{sec:setup}
% INSPIRED FROM https://ieeexplore.ieee.org/stamp/stamp.jsp?tp=&arnumber=9831435&tag=1

\subsection{Problem Statement}
\label{sec:problem_statement}
The objective of this study is to identify which devices are currently active.
The total electrical power $p$ consumed by a household at any given moment $t$ is calculated as the sum of the power used by each electrical device, denoted as $p_i(t)$, where there are $N_{d}$ devices in total as defined in Eq.~\ref{eqn:problem1}. Additionally, measurement noise (including any unidentified residual devices) $e(t)$ is also taken into account. The status indicator $s(t)$ determines the activity of each device, where $s(t) = 0$ indicates that the device is inactive and $s(t) = 1$ indicates that the device is active at the given moment $t$.
\begin{equation}
\label{eqn:problem1}
p(t) = \sum_{i=1}^{N_{d}} s_i(t) p_i(t) + e(t)
\end{equation}

To solve the problem and thus estimate the status indicator $s_i(t)$ for each device, we can employ classical ML or DL for multi-label classification of devices. Devices are classified as active if the corresponding status indicator $s_i(t)$ predicted by the model exceeds 0.5, as illustrated in Figure~\ref{fig:problem}. The cardinality of the set $s$ representing all the possible active devices, denoted by $|s|$, indicates the number of labels that need to be recognized. In the context of this paper, the value of $s$ varies between experiments as explained in Sections~\ref{subsec:testA} and~\ref{subsec:testB}.

\begin{figure}
    \centering
    \subfloat[REFIT]{
    \includegraphics[width=0.9\linewidth, clip, trim={0 0 0 0}]{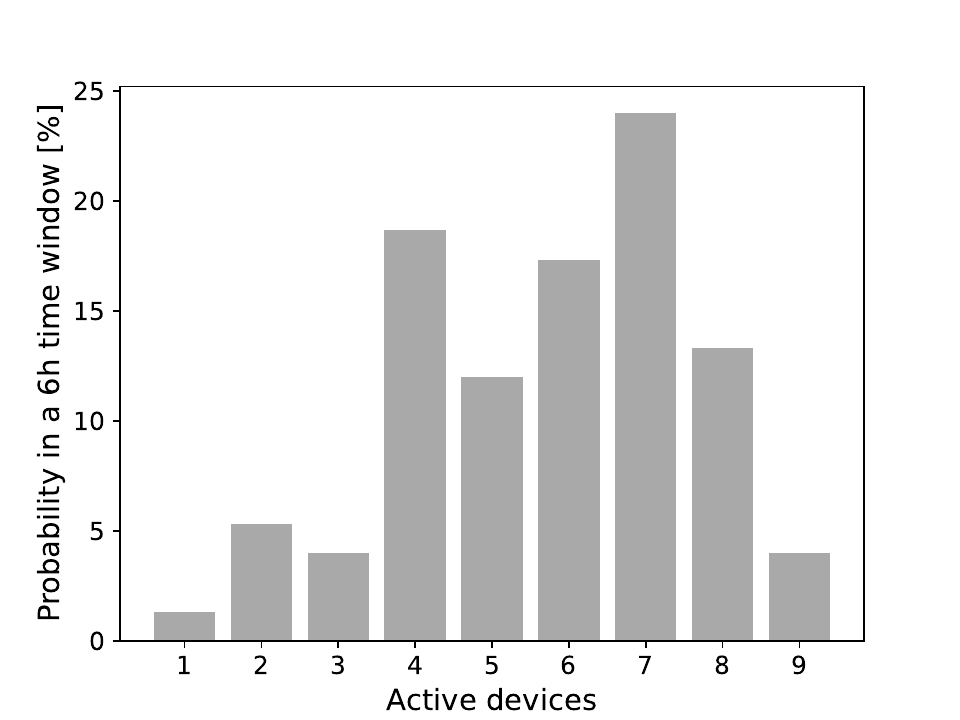}
    }\\
    \subfloat[UK-DALE]{
    \includegraphics[width=0.9\linewidth, clip, trim={0 0 0 0}]{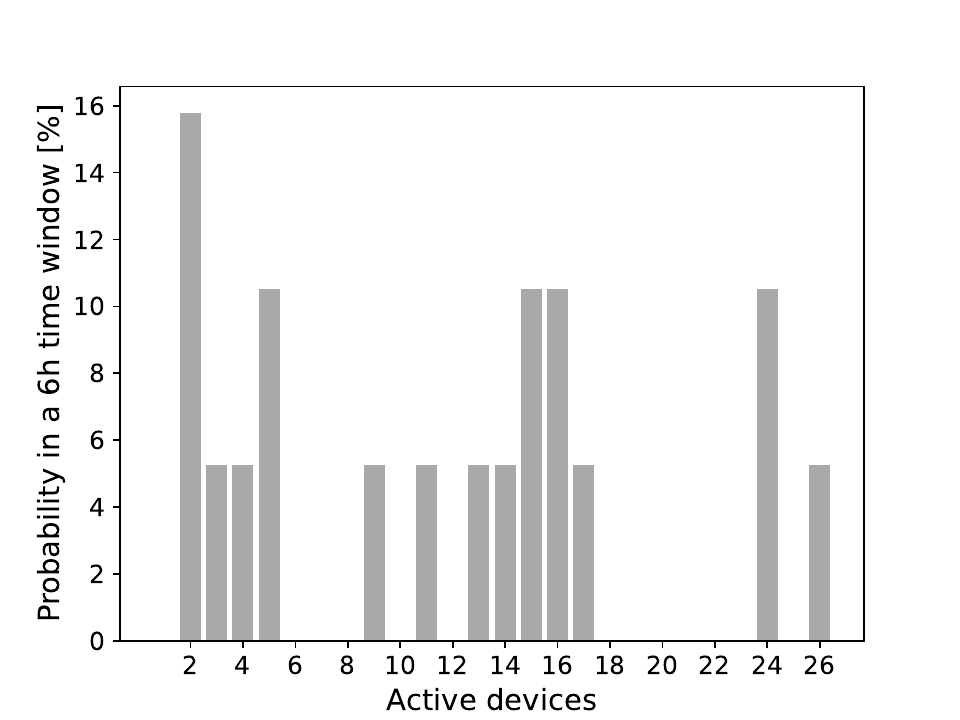}
    }
    \caption{Probability distribution of the number of active devices in a 6 hours time window in REFIT and UK-DALE datasets.}
    \label{fig:datasetanalysis}
\end{figure}
\begin{figure*}[ht]
    \centering
    \includegraphics[width=1.0\linewidth, clip, trim={0 0 0 0}]{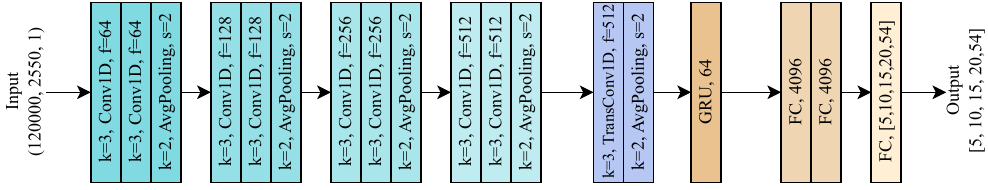}
    \caption{The proposed \gls{ctrnn} architecture inspired by the VGG family of architectures is explained within the figure, where "k" signifies the kernel, "f" represents the number of filters, and "s" denotes the stride value.}
    \label{fig:model}
\end{figure*}

\subsection{Methodology}
\label{sec:methodology}

To this end, this work examines a novel methodology depicted in Figure~\ref{fig:AB-Explanation} that more accurately represents the performance of models in practical use cases. We approached the problem by first analysing the dataset to identify the maximum number of active devices $N_{max.AD}$ in the time windows that the model is trained on and the average maximum number $\overline{N_{max.AD}}$. We then generated two distinct groups of mixed datasets, each comprising different sets of active and inactive devices. The first group contained mixed datasets with a fixed number of $N_{max.AD}$ active devices, whereas the second group included mixed datasets with varying number of active devices between 1 and $\overline{N_{max.AD}}$. To generate the groups of mixed datasets we used REFIT~\cite{Murray2017refit} and UK-DALE~\cite{kelly2015uk-dale} low-frequency datasets, also present in Tanoni~\textit{et al.}~\cite{Tanoni2022} and Langevin~\textit{et al.}~\cite{Langevin2022}. 

We propose a methodology\footnote{https://github.com/anzepirnat/CtRNN/} that aims to assess classical ML and DL models in a realistic scenario, which differs from the approaches commonly taken in recent works \cite{Tanoni2022, Pan2020, D'Incecco2020, Wang2022, Zhang2018} and \cite{Kelly2015} that only use 5 distinct devices for evaluation. This limited number does not represent the typical diversity of devices encountered in real-world settings. For instance, the analysis of REFIT and UK-DALE datasets in Figure~\ref{fig:datasetanalysis} reveals the presence of up to 9 and 26 active devices, respectively. Therefore, our methodology considers a wider range of devices for a more accurate evaluation of DL models in realistic conditions as depicted.

\subsubsection{Sensitivity Evaluation Group} %NAME
\label{subsec:testA}

The Sensitivity Evaluation (SE) is a group of multiple mixed datasets that cover cases where there are 5, 10, 15 or 20 DiT in the household and 1, 2, ..., $N_{max.AD}$ of them are AD. The number of DiT is used universally across all datasets and the maximum number of AD ($N_{max.AD}$) is used as a parameter depending on the maximum number of active devices in the time window that we are training our model on. 

Mixed datasets in SE Group provide an insight into how the model in testing performs depending on the number of AD in four general cases of DiT, i.e. 5, 10, 15 and 20. In case of the UK-DALE dataset we also give an insight into a case of 54 DiT, which significantly exceeds the maximum number of DiT in the REFIT dataset. 

In our case we were using 2550 samples for training, with sample rate of REFIT and UK-DALE that results in approximately 6 hour long time window. In the given time window there is a number of AD ranging from 1 to 9 for REFIT and from 2 to 26 for UK-DALE, as shown in Figure~\ref{fig:datasetanalysis}, thus $N_{max.AD} = 9$ for REFIT and $N_{max.AD} = 26$ for UK-DALE.

\subsubsection{Realistic Evaluation Group} %NAME
\label{subsec:testB}

The Realistic Evaluation Group (RE) is an extension of the methodology employed in many recent works \cite{Tanoni2022, Langevin2022, Pan2020, D'Incecco2020, Wang2022, Zhang2018, Kelly2015} which account for only 5 distinct devices. We propose a group of multiple mixed datasets that cover cases where there are 5 and also 10, 15 and 20 devices in total (DiT) in the household and chosen at random. We generate mixed datasets with an equal mix of all possible numbers of ADs. Thus, we generated 4 mixed datasets, each containing samples with 1, 2, ..., $\overline{N_{max.AD}}$ ADs. Such RE Group presents more practical evaluation of the model by simulating a real-world scenario in which households utilize a variety of active devices rather than a fixed number.
 
In our case the average maximum number ($\overline{N_{max.AD}}$) of active devices was 8 for REFIT and 14 for UK-DALE as supported by the results in Figure~\ref{fig:datasetanalysis}.  
Therefore, the training data comprised time windows with ADs ranging from 1 to 8 and 1 to 14 for REFIT and UK-DALE, respectively. However, in cases where there were fewer devices in the household than active devices, the range of ADs was set to 1 to \textit{DiT-1}. For instance, when there were 5 DiT in the household, the range of ADs was from 1 to 4, as was the case for both REFIT and UK-DALE datasets. Similarly, when there were 10 devices in the household, the range was from 1 to 9 ADs. Lastly it should be noted that we used an 80:20 split for training and evaluation parts of all datasets in this research.

\section{Proposed Neural Network Architecture}
\label{sec:model}

To solve the problem defined in Section~\ref{sec:problem_statement} we introduce the  novel \gls{ctrnn} architecture based on the VGG family. Architectures from this family, adapted for time series data, have  previously proved successful in NILM disaggregation tasks~\cite{Perez2021VGGinNILM}. Additionally, the hyper-parameters of the architectures were determined empirically following the principles derived from our prior work \cite{Pirnat2022}, determining the ratio between prediction performance and computational complexity. The computational complexity of a network, measured in floating-point operations (FLOPs), is determined by the number of layers $L$ and the computational complexity of each separate layer $F_\text{l}$, as described in  Eq.~\ref{eqn:ModelFlops}. Throughout our empirical design phase, we explored networks with layers ranging from $L\in\{16, ..., 22\}$.
	
\begin{equation}
	\label{eqn:ModelFlops}
	N_{FLOPs} = \sum_{l=1}^{L} F_\text{l} ,
\end{equation}

In addition to the VGG adaptation, our architecture also includes the transposed convolutional layer (TCNN) and the gated recurrent unit (GRU) layers. The transposed convolutional layer increases the temporal resolution of features while reducing the number of features from the previous layer~\cite{Zhou2022}, while GRU is utilised to better capture the temporal correlation in the TS and showed great potential in solving NILM related tasks~\cite{Tanoni2022, Langevin2022}. The combination of CNNs, TCNNs and GRU layers enable the architecture to better capture the spatial and temporal correlation within the time series data.

The resulting architecture is illustrated in Figure~\ref{fig:model}, where each layer is depicted with its type and hyper-parameters. The architecture comprises four blocks, each consisting of two convolutional layers and one average pooling layer. The number of filters in each block doubles, starting from 64. Following the convolutional blocks, there is a TCNN layer and a GRU layer. Prior to the output layer, there are two fully-connected layers with 4096 nodes. The number of nodes in the output layer is adjusted to meet the specific requirements and ranges from 5 to 54, depending on the used  dataset. All layers utilize the ReLU activation function, except for the output layer, which employs the sigmoid activation function.

\subsection{Computation Efficiency Considerations}
\label{subsec:ComputationEfficiencyConsiderations}
As the purpose of using NILM is to reduce energy consumption, it is logical to ensure that the process itself is as energy-efficient as possible. Thus, our goal was to design a deep learning architecture that surpasses the state of the art not only in performance but also in terms of energy efficiency.

In order to assess the energy consumption of the architecture, it is necessary to calculate its complexity. This typically involves adding up the total number of FLOPs required for each layer.

We estimate the complexity of the most energy-consuming layers, namely the convolutional, pooling, and fully-connected layers, using the equations presented by Pirnat~\textit{et al.} in~\cite{Pirnat2022}. In addition, we calculate the complexity of the GRU layer using the equation proposed in~\cite{Zhang2018-2}.

We use equations from~\cite{Pirnat2022} to estimate the energy consumption of the proposed architecture and for comparison also for one popular reference architecture from the VGG family of architectures, i.e., VGG11 \cite{simonyan2015very}, and for the two architectures used as a reference in performance evaluation, i.e., TanoniCRNN~\cite{Tanoni2022} and VAE-NILM~\cite{Langevin2022}. This is done with an assumption that the architecture is trained and used on an Nvidia A100 graphics card and that each kWh of electricity produced results in 250g of CO$_2$ equivalent emissions (as it is the case for Slovenia).

\begin{table*}[htbp]
\ra{1.3}
\caption{Table of training parameters for CtRNN and VGG11, for SE Group mixed datasets derived from REFIT and UK-DALE (BS - batch size; LR - learning rate; E - no. of epochs).}
\label{tab:TestATrainingParameters}
\centering 
%\begin{tabular}{p{1.5cm} p{4cm} p{1.5cm}}
\renewcommand{\arraystretch}{0.9}
\resizebox{\textwidth}{!}{%
\begin{tabular}{@{}l|rcr|rcr|rcr|rcr|rcr|rcr|rcr@{}}
    \toprule
    %\footnotesize
    
    %|||||||
    % CtRNN
    %|||||||

    %-------
    % REFIT
    %-------
    \multirow{2}{*}[-0.5em]{\textbf{CtRNN REFIT}}
    & \multicolumn{3}{c|}{\textbf{1 AD}}
    & \multicolumn{3}{c|}{\textbf{2 AD}}
    & \multicolumn{3}{c|}{\textbf{3, 4 AD}}
    & \multicolumn{3}{c|}{\textbf{5, 6, 7, 8 AD}}
    & \multicolumn{3}{c|}{\textbf{9 AD}}
    \\
    \cmidrule(lr){2-4}
    \cmidrule(lr){5-7}
    \cmidrule(lr){8-10}
    \cmidrule(lr){11-13}
    \cmidrule(l){14-16}

    {}
    & \textbf{BS} & \textbf{LR} & \textbf{E} 
    & \textbf{BS} & \textbf{LR} & \textbf{E}
    & \textbf{BS} & \textbf{LR} & \textbf{E}
    & \textbf{BS} & \textbf{LR} & \textbf{E}
    & \textbf{BS} & \textbf{LR} & \textbf{E}
    \\
    %\midrule
    \cline{1-16}
    5 DiT
    & 512 & $10^{-4}$ & 40 
    & 256 & 5$\times$$10^{-4}$ & 20
    & 128 & 5$\times$$10^{-4}$ & 20
    & / & / & /
    & / & / & /
    \\
    10 DiT
    & 512 & $10^{-4}$ & 40 
    & 256 & 5$\times$$10^{-4}$ & 20
    & 128 & 5$\times$$10^{-4}$ & 20
    & 128 & 5$\times$$10^{-4}$ & 20
    & 128 & $10^{-4}$ & 20
    \\
    15 DiT
    & 512 & $10^{-4}$ & 40 
    & 256 & 5$\times$$10^{-4}$ & 20
    & 128 & 5$\times$$10^{-4}$ & 20
    & 128 & 5$\times$$10^{-4}$ & 20
    & 128 & 5$\times$$10^{-4}$ & 20
    \\
    20 DiT
    & 512 & $10^{-4}$ & 40 
    & 256 & 5$\times$$10^{-4}$ & 20
    & 128 & 5$\times$$10^{-4}$ & 20
    & 128 & 5$\times$$10^{-4}$ & 20
    & 128 & 5$\times$$10^{-4}$ & 20
    \\
    %-------
    % UK-DALE
    %-------
    %\midrule
    %\cline{1-19}
    \midrule
    \multirow{2}{*}[-0.5em]{\textbf{CtRNN UK-DALE}}
    & \multicolumn{3}{c|}{\textbf{2, 3, 4 AD}}
    & \multicolumn{3}{c|}{\textbf{5, 9 AD}}
    & \multicolumn{3}{c|}{\textbf{11, 13 AD}}
    & \multicolumn{3}{c|}{\textbf{14 AD}}
    & \multicolumn{3}{c|}{\textbf{15, 16, 17 AD}}
    & \multicolumn{3}{c|}{\textbf{24, 26 AD}}
    \\
    \cmidrule(lr){2-4}
    \cmidrule(lr){5-7}
    \cmidrule(lr){8-10}
    \cmidrule(lr){11-13}
    \cmidrule(lr){14-16}
    \cmidrule(l){17-19}
    {}
    & \textbf{BS} & \textbf{LR} & \textbf{E} 
    & \textbf{BS} & \textbf{LR} & \textbf{E}
    & \textbf{BS} & \textbf{LR} & \textbf{E}
    & \textbf{BS} & \textbf{LR} & \textbf{E}
    & \textbf{BS} & \textbf{LR} & \textbf{E}
    & \textbf{BS} & \textbf{LR} & \textbf{E}
    \\
    \midrule
    5  DiT
    & 128 & 3$\times$$10^{-4}$ & 20 
    & / & / & /
    & / & / & /
    & / & / & /
    & / & / & /
    & / & / & /
    \\
    10 DiT
    & 128 & 3$\times$$10^{-4}$ & 20 
    & 128 & 3$\times$$10^{-4}$ & 20
    & / & / & /
    & / & / & /
    & / & / & /
    & / & / & /
    \\
    15 DiT
    & 128 & 3$\times$$10^{-4}$ & 20 
    & 128 & 3$\times$$10^{-4}$ & 20
    & 128 & 5$\times$$10^{-4}$ & 20
    & 128 & 5$\times$$10^{-4}$ & 20
    & / & / & /
    & / & / & /
    \\
    20 DiT
    & 128 & 3$\times$$10^{-4}$ & 20 
    & 128 & 3$\times$$10^{-4}$ & 20
    & 128 & 5$\times$$10^{-4}$ & 20
    & 128 & $10^{-4}$ & 20
    & 128 & $10^{-4}$ & 20
    & / & / & /
    \\
    54 DiT
    & 128 & $10^{-4}$ & 20 
    & 128 & $10^{-4}$ & 20
    & 128 & 3$\times$$10^{-4}$ & 20
    & 128 & 3$\times$$10^{-4}$ & 20
    & 128 & 3$\times$$10^{-4}$ & 20
    & 128 & 3$\times$$10^{-4}$ & 20
    \\
    
    \midrule
    \midrule
    \midrule

    %|||||||
    % VGG11
    %|||||||
    
    %-------
    % REFIT
    %-------
    \multirow{2}{*}[-0.5em]{\textbf{VGG11 REFIT}}
    & \multicolumn{3}{c|}{\textbf{1 AD}}
    & \multicolumn{3}{c|}{\textbf{2 AD}}
    & \multicolumn{3}{c|}{\textbf{3, 4 AD}}
    & \multicolumn{3}{c|}{\textbf{5, 6, 7, 8, 9 AD}}
    \\
    \cmidrule(lr){2-4}
    \cmidrule(lr){5-7}
    \cmidrule(lr){8-10}
    \cmidrule(l){11-13}
    {}
    & \textbf{BS} & \textbf{LR} & \textbf{E} 
    & \textbf{BS} & \textbf{LR} & \textbf{E}
    & \textbf{BS} & \textbf{LR} & \textbf{E}
    & \textbf{BS} & \textbf{LR} & \textbf{E}
    \\
    \cline{1-13}
    5  DiT
    & 512 & $10^{-4}$ & 50 
    & 256 & $10^{-4}$ & 20
    & 128 & $10^{-4}$ & 20
    & / & / & /
    \\
    10 DiT
    & 512 & $10^{-4}$ & 50 
    & 256 & $10^{-4}$ & 20
    & 128 & $10^{-4}$ & 20
    & 128 & $10^{-4}$ & 20
    \\
    15 DiT
    & 512 & $10^{-4}$ & 50 
    & 256 & $10^{-4}$ & 20
    & 128 & $10^{-4}$ & 20
    & 128 & $10^{-4}$ & 20
    \\
    20 DiT
    & 512 & $10^{-4}$ & 50 
    & 256 & $10^{-4}$ & 20
    & 128 & $10^{-4}$ & 20
    & 128 & $10^{-4}$ & 20
    \\
    
    %-------
    % UK-DALE
    %-------
    %\cline{1-16}
    \cmidrule(){1-16}
    %\cline{1-16}
    \multirow{2}{*}[-0.5em]{\textbf{VGG11 UK-DALE}}
    & \multicolumn{3}{c|}{\textbf{2, 3, 4 AD}}
    & \multicolumn{3}{c|}{\textbf{5, 9 AD}}
    & \multicolumn{3}{c|}{\textbf{11, 13, 14 AD}}
    & \multicolumn{3}{c|}{\textbf{16, 17}}
    & \multicolumn{3}{c|}{\textbf{24, 26 AD}}
    \\
    \cmidrule(lr){2-4}
    \cmidrule(lr){5-7}
    \cmidrule(lr){8-10}
    \cmidrule(lr){11-13}
    \cmidrule(l){14-16}

    {}
    & \textbf{BS} & \textbf{LR} & \textbf{E} 
    & \textbf{BS} & \textbf{LR} & \textbf{E}
    & \textbf{BS} & \textbf{LR} & \textbf{E}
    & \textbf{BS} & \textbf{LR} & \textbf{E}
    & \textbf{BS} & \textbf{LR} & \textbf{E}

    \\
    \cline{1-16}
    %\cmidrule(lr){1-16}
    %\midrule
    5  DiT
    & 128 & $10^{-4}$ & 20 
    & / & / & /
    & / & / & /
    & / & / & /
    & / & / & /
    \\
    10 DiT
    & 128 & $10^{-4}$ & 20 
    & 128 & $10^{-4}$ & 20
    & / & / & /
    & / & / & /
    & / & / & /
    \\
    15 DiT
    & 128 & $10^{-4}$ & 20 
    & 128 & $10^{-4}$ & 20
    & 128 & $10^{-4}$ & 20 
    & / & / & /
    & / & / & /
    \\
    20 DiT
    & 128 & $10^{-4}$ & 20 
    & 128 & $10^{-4}$ & 20
    & 128 & $10^{-4}$ & 20 
    & 128 & 5$\times$$10^{-5}$ & 20
    & / & / & /
    \\
    54 DiT
    & 128 & $10^{-4}$ & 20 
    & 128 & $10^{-4}$ & 20
    & 128 & $10^{-4}$ & 20 
    & 128 & 5$\times$$10^{-5}$ & 20
    & 128 & 5$\times$$10^{-5}$ & 20
    \\
\bottomrule
\end{tabular}
}
\end{table*}

\subsection{Evaluation Datasets and Training Parameters}
\label{subsec:tests}

We first compared the performance of our model to VGG11~\cite{simonyan2015very} and with the performance of the model created by Tanoni~\textit{et al.}~\cite{Tanoni2022} adapted to fully supervised DL, and to results achieved by Langevin~\textit{et al.}~\cite{Langevin2022}. This comparison was done on the standard evaluation methodology defined in \cite{Tanoni2022} that comprises a total of 5 distinct devices: fridge, washing machine, dish washer, microwave and kettle. Those 5 devices were also used by many recent works \cite{Tanoni2022, Langevin2022, Pan2020, D'Incecco2020, Wang2022, Zhang2018, Kelly2015} where they didn't exactly specify the number of active devices, thus we chose to reproduce the one in \cite{Tanoni2022}. Samples with varying numbers of active devices from 1 to 4 are randomly interspersed throughout the mixed dataset. To make the training and evalution parts of the dataset we used an 80:20 split.  
%\begin{figure*}[htb]
%    \centering
%    \includegraphics[width=0.7\linewidth]{figs/datasets4.pdf}
%    \caption{The proposed groups of datasets A and B}
%    \label{fig:datasets}
%\end{figure*}

For this comparison we used a learning rate of 0.0003 and 20 epochs for our model while for the TanoniCRNN model we adopted the parameters specified as optimal in~\cite{Tanoni2022}, which include the same number of epochs and a different learning rate of 0.002. Moreover, we used the same batch size of 128 for both models.

Subsequently we compared the performance of our model with that of VGG11 and RF on the two groups of mixed datasets described in Section~\ref{sec:methodology}. We choose VGG11 as a benchmark because VGG architectures are adopted in recent works \cite{Weicong2020VGGinNILM, Dogsheng2020VGGinNILM} for classification in NILM due to their effectiveness, and VGG11 is the closest match regarding the complexity. RF was chosen as the benchmark as it was reported to be the best classic ML algorithm for ON/OFF classification on NILM data in a previous study~\cite{Wu2019}.

Training parameters used for SE Group are presented in Table~\ref{tab:TestATrainingParameters} for CtRNN and VGG11 as follows. Each sub-table includes information about the parameters corresponding to that combination of model and dataset, the first one addressing CtRNN with REFIT, the second concerning CtRNN and  UK-DALE, while the third and fourth describing VGG11 with REFIT and UK-DALE respectively. For each model-dataset combination, parameters for training data with various DiT, namely 5, 10, ..54, for the SE group described in Section \ref{subsec:testA} are provided. The grouped columns include the parameter values of the architectures, namely the BS, LR and E. BS denotes the batch size, representing the total number of training packets that are sent to the architecture at a time. LR stands for learning rate, which is a parameter that determines the step size at each iteration of the architecture. Finally, E signifies the number of epochs, one epoch representing one pass of training data through the architecture. As can be seen in the table, the three parameters differ across the architecture-dataset, DiT and AD combinations. For instance, for CtRNN with REFIT, 5 DiT and 1 AD, the BS is $512$, LR is $10^{-4}$ and E is $40$.

In summary, the batch size used for training both models was predominantly set to 128, with some variations of 256 or 512. The epoch count was set to 20 for both models in most cases, but it was also set to 50 for VGG11 and 40 for CtRNN, respectively, in certain scenarios, because they benefited from more training passes. The learning rate ranged changed between $10^{-4}$ and $5\times10^{-5}$ for VGG11 and between $5\times10^{-4}$ and $5\times10^{-5}$ for CtRNN, since both required a larger or smaller step size in certain scenarios. The batch size and the number of epochs were similar to ~\cite{Tanoni2022}, and fine tuned through an empirical process. The learning rate was also empirically tuned.

The training parameters used for RE Group had less variation, always using batch size 128 and 20 epochs, therefore the replication of the study can be done with the numbers provided in this paragraph. The learning rate for CtRNN was 0.0003 on both REFIT and UK-DALE datasets; for VGG11 it was 0.0001 on both REFIT and UK-DALE datasets. In all tests the batch size used is selected as the largest we could run or the one that gives the best results, selected through trial and error. It is equal for both models in the test as performance vary only slightly depending on its size. 

\subsection{Metrics}
\label{subsec:metrics}

We evaluate the performance with a combined metric average weighted F1 score ($\overline{F1score_{w}}$), %Further explanation on why we use average weighted F1 score and not just average F1 score. 
since performance evaluation based on a simple arithmetic mean of the F1 score would fail to provide an accurate reflection of the overall performance because our mixed datasets are generated in a way that does not provide each device equal representation. The use of weights ensures that all devices will affect the average score proportionally to how often they appear in the particular mixed dataset.

\begin{equation}
\label{eqn:F1weighted}
\overline{F1score_{w}} = \sum_{i=1}^{N_{d}} F1score_{i} \times Weight_{i}
\end{equation}
Average weighted F1 score is based on three metrics: true positive (TP), false positive (FP), and false negative (FN). TP represents the cases where the device is correctly classified as active, FP represents the cases where the device is incorrectly classified as active, and FN represents the cases where the device is incorrectly classified as inactive.

Using these metrics, we calculate the precision $Precision = \frac{TP}{TP + FP}$ and recall $Recall = \frac{TP}{TP + FN}$, and from these, we derive the F1 score $F1score = 2 \times \frac{Precision \times Recall}{Precision + Recall}$. To obtain the average weighted F1 score defined in Eq.~\ref{eqn:F1weighted}, we calculate the F1 score for each device and then take the average based on their weight $Weight = \frac{SSD}{SAD}$, which is determined by the support for the specified device (SSD) and the support of all devices (SAD).

\section{Performance Evaluation}
\label{sec:evaluation}
Using the proposed methodology with two groups of mixed datasets, we carried out comprehensive performance evaluation of CtRNN DL architecture and benchmarked it against selected state-of-the-art architectures in terms of energy efficiency and accuracy of determining the status of devices, as described in the following.

\subsection{Energy Consumption}
\label{subsec:EECO2}
\begin{table}[htb]
    \ra{1.3}
    \caption{Energy used in training the proposed CtRNN model in comparison to VGG11, TanoniCRNN and VAE-NILM.}%, for each mixed dataset in group B and group A where batch size is 128}
    \label{tab:CarbonFootprint1}
    \centering
    \resizebox{0.95\columnwidth}{!}{
    \begin{tabular}{l|c|c|c}%|c}
        \toprule
        \bfseries NN & \bfseries parameters & \bfseries FLOPs  & \bfseries energy \\%& \bfseries carbon footprint \\
        \midrule
        CtRNN
        & 19.6\,$\cdot\,10^6$
        & 0.85\,$\cdot\,10^9$
        & \textbf{1.51\,MJ}        
        %& \textbf{104.9\,g\,CO$_2$\,eq.}
        \\\midrule
        VGG11~\cite{simonyan2015very} 
        & 185.6\,$\cdot\,10^6$
        & 1.21\,$\cdot\,10^9$ 
        & 2.15\,MJ
        %& 149.3\,g\,CO$_2$\,eq. 
        \\
        TanoniCRNN~\cite{Tanoni2022} 
        & \textbf{0.75}\boldmath{\,$\cdot\,10^6$}
        & 1.11\,$\cdot\,10^9$
        & 1.97\,MJ
        %& 136.7\,g\,CO$_2$\,eq.
        \\
        VAE-NILM~\cite{Langevin2022}
        & 3.8\,$\cdot\,10^6$
        & \textbf{0.42}\boldmath{\,$\cdot\,10^9$}
        & 13.2\,-\,263\,MJ$^{*}$ % 3.7 - 73.1 kwh
        %& 0.93\,-\,18.3\,kg\,CO$_2$\,eq.
        \\
    \bottomrule
    \end{tabular}
    }
    
    \footnotesize{\textsuperscript{*} For VAE-NILM,  it is only possible to compute a range of values based on the reported epochs ~\cite{Langevin2022}  between  5 and 100.}
    
\end{table} 
\begin{figure}
  \centering
  \includegraphics[width=0.5\textwidth]{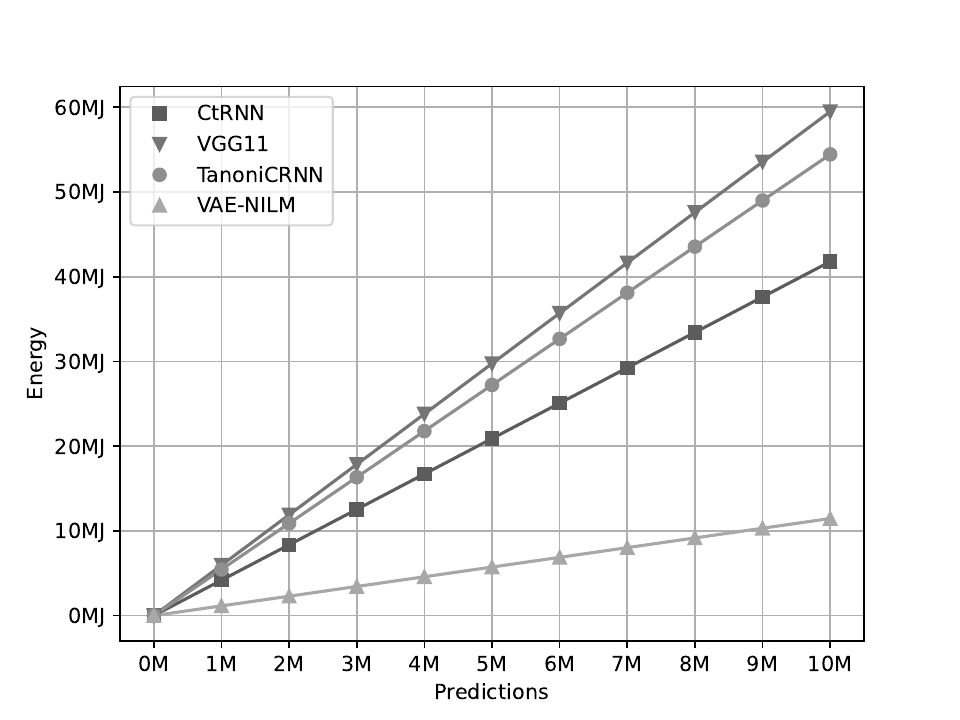}
  \caption{Energy used for making predictions with the proposed model in comparison to VGG11, TanoniCRNN and VAE-NILM.}
  \label{fig:EnergyPredictions}
\end{figure}

\begin{table*}[htb]
\ra{1.3}
\caption{Average weighted F1 score results for CtRNN compared to TanoniCRNN~\cite{Tanoni2022} and VAE-NILM~\cite{Langevin2022} on mixed dataset of 5 devices derived from REFIT and UK-DALE.}
\label{tab:LOL2tanonicompRefitAndUk-dale}
\centering
\resizebox{0.8\textwidth}{!}{%
\begin{tabular}{@{}l||c|c|c||c|c|c@{}}
    \toprule

    %devices
    \multirow{2}{*}[-0.5em]{\textbf{devices}}
    & \multicolumn{3}{c||}{\textbf{REFIT}}
    & \multicolumn{3}{c}{\textbf{UK-DALE}}
    %& \multicolumn{3}{c|}{3, 4 AD}
    %& \multicolumn{3}{c|}{5, 6, 7, 8, 9 AD}
    \\
    \cmidrule(lr){2-4}
    \cmidrule(lr){5-7}
    %\cmidrule(lr){8-10}
    %\cmidrule(l){11-13}
    %{}
    %& BS & LR & E 
    %& BS & LR & E
    %& BS & LR & E
    %& BS & LR & E
    %\\
    & \textbf{CtRNN}
    & \textbf{TanoniCRNN~\cite{Tanoni2022}}
    & \textbf{VAE-NILM~\cite{Langevin2022}}
    & \textbf{CtRNN}
    & \textbf{TanoniCRNN~\cite{Tanoni2022}}
    & \textbf{VAE-NILM~\cite{Langevin2022}}
    \\
    \midrule
    fridge
    %LOL2
    & \textbf{0,93}
    %tanoni
    & 0,92
    %Langevin
    & 0,85
% We merge table together:
    %LOL2
    & \textbf{1,0}
    %tanoni
    & \textbf{1,0}
    %Langevin
    & 0,81    
    \\
    washing machine
    %LOL2
    & \textbf{0,89}
    %tanoni
    & 0,84
    %Langevin
    & 0,78
% We merge table together:
    %LOL2
    & \textbf{0,92}
    %tanoni
    & 0,81
    %Langevin
    & 0,74
    \\
    dish washer
    %LOL2
    & \textbf{0,88}
    %tanoni
    & 0,80
    %Langevin
    & 0,84
% We merge table together:
    %LOL2
    & \textbf{0,87}
    %tanoni
    & 0,86
    %Langevin
    & 0,65
    \\
    microwave
    %LOL2
    & \textbf{0,89}
    %tanoni
    & 0,71
    %Langevin
    & 0,59
% We merge table together:
    %LOL2
    & \textbf{0,96}
    %tanoni
    & 0,80
    %Langevin
    & 0,32
    \\
    kettle
    %LOL2
    & \textbf{0,95}
    %tanoni
    & 0,87
    %Langevin
    & 0,87
% We merge table together:
    %LOL2
    & \textbf{0,93}
    %tanoni
    & 0,86
    %Langevin
    & 0,87
    \\
    \midrule
    weighted avg
    %LOL2
    & \textbf{0,91}
    %tanoni
    & 0,83
    %Langevin
    & 0,78
% We merge table together:
    %LOL2
    & \textbf{0,94}
    %tanoni
    & 0,87
    %Langevin
    & 0,68
    \\
\bottomrule
\end{tabular}
}
\end{table*}

The results of our energy consumption evaluation, performed according to the considerations in Section~\ref{subsubsec:CO2} and methodology described in Section~\ref{subsec:tests}, for training different architectures for each mixed dataset from RE Group or SE Group using batch size 128 are displayed in Table~\ref{tab:CarbonFootprint1}. The rows of the table list the neural-network arhitectures considered in this work, namely the proposed CtRNN and the VGG, TanoniCRNN and VAE-NILM baselines selected in Section \ref{sec:model}. The first column of the table displays the number of internal parameters for each of the NNs, that represent the total number of weights and biases in the NNs. The second column lists the number of FLOPs, which is the number of floating point operations needed for a pass through the NN. The third column showcases energy consumed during the training of the models. The values in second and third columns are calculated as explained in Section~\ref{subsec:ComputationEfficiencyConsiderations}. Despite TanoniCRNN having the lowest number of parameters, this does not result in lowest number of FLOPs, energy consumption. Similarly, although VAE-NILM exhibits the lowest number of FLOPs, it has the highest energy consumption. These factors are clearly influenced by additional training parameters, such as the number of epochs and batch size required to achieve satisfactory results. The proposed model outperforms state-of-the-art TanoniCRNN in terms of energy consumption, with 23.3\,\% less energy consumed on a mixed dataset tested on five commonly used devices. In addition, compared to VGG11 on groups A and B, the proposed model demonstrates 29.7\,\% less energy consumed.%]

% Text about the energy predictions graph

% 1) you introduce the figure
In addition to the energy used during the training, energy consumed for making predictions can also be significant when the number of requests for predictions is high as depicted in Figure~\ref{fig:EnergyPredictions}. 
% 2) you explain what's in it by describing the axes
On the x-axis the figure plots the number of predictions from 0 to 10 million while on the y-axis it plots the consumed energy in mega Joules. 
The results show that in making 10M predictions our model consumes 41.8MJ, while TanoniCRNN, VGG11 and VAE-NILM consume: 54.4MJ, 59.5MJ and 11.5MJ. VAE-NILM consumes notably less energy then other models, that can be attributed to Langevin et al.~\cite{Langevin2022} using a window of 1024 samples for training, while we used 2550 for all others, and because it has less FLOPs. 
The figure also shows that for more than a million predictions, the energy consumed exceeds the energy used for training the models, with exception of VAE-NILM for which only a range can be computed based on the number of epoch provided in ~\cite{Langevin2022}.

\subsection{Results on Mixed Dataset with 5 Commonly Used Devices}
\label{subsubsec:TanoniComparison}
Comparison with Tanoni~\cite{Tanoni2022} and Langevin~\cite{Langevin2022} on datasets of 5 devices derived from the REFIT and UK-DALE demonstrates superior performance of our proposed model with a significant gap in F1 score as can be seen in Table~\ref{tab:LOL2tanonicompRefitAndUk-dale}. The first column of the table lists the devices, columns 2-4 lists F1 scores for the models trained on the REFIT dataset while columns 5-7 on the ones trained on the UK-DALE dataset. For each of the datasets, the subcolumns list the evaluated architectures, namely CtRNN, TanoniCRNN and VAE-NILM. Rows 2-6 present per device F1 scores while row 7 provides the weighted average of the F1 score. It can be seen from the table that the proposed model displays superior performance on all devices on both datasets with the exception of fridge in UK-DALE where its even with TanoniCRNN. That is also evident from the average weighted F1-score in the final row. Specifically, on the REFIT derived dataset, our model achieves an average weighted F1 score of 91\,\% compared to 83\,\% and 78\,\% obtained by TanoniCRNN and VAE-NILM, respectively, an improvement of 8 and 13 percentage points. On the UK-DALE derived dataset, our model outperforms TanoniCRNN and VAE-NILM by 7 and 26 percentage points, respectively, achieving an average weighted F1 score of 94\,\% compared to their 87\,\% and 68\,\%. 

A closer analysis of the two best models from Table~\ref{tab:LOL2tanonicompRefitAndUk-dale} shows, according to the second row of the table, that our approach slightly outperforms the approach of \cite{Tanoni2022} in recognizing the fridge class with an F1 of 0.93 compared to 0.92 on the REFIT dataset, while both approaches work perfectly on the UK-DALE dataset. However, the fridge class of devices is easier to identify as its consumption pattern is periodic and is the most pronounced of all appliances. The real difference in performance between the two models is seen in the detection of appliances with short consumption intervals. In row 3 of Table~\ref{tab:LOL2tanonicompRefitAndUk-dale}, our method outperforms TanoniCRNN~\cite{Tanoni2022} in detecting washing machines on the REFIT dataset by 5 percentage points, and is 11 percentage points better on the UK-DALE dataset. When detecting dishwashers in row 4, our method is 8 percentage points superior  than \cite{Tanoni2022} on the REFIT dataset, while it is 1 percentage point better on the UK-DALE dataset. The largest difference in performance can be observed in row 5 for the microwave class. Here, our method with an F1 of 0.89 is significantly superior compared to  \cite{Tanoni2022}, with an F1 of 0.71 on the REFIT dataset. Something similar can be observed on the UK-DALE dataset, where our method achieves an F1 score of 0.96 compared to the F1 score of 0.80 for \cite{Tanoni2022}. If we look at the kettle class in row 6 of Table~\ref{tab:LOL2tanonicompRefitAndUk-dale}, we can again see that our method outperforms \cite{Tanoni2022} by 8 and 7 percentage points in both the REFIT and UK-DALE datasets, respectively.

These results show that our proposed architecture is significantly better in detecting appliances with shorter consumption duration, compared to \cite{Tanoni2022}. The reason for that is, that our architecture design is superior at detecting both spatial and temporal correlation within the signal. Spatial correlations are detected by the convolutional layers while temporal by the GRU layers of the architecture described in Section \ref{sec:model} and depicted in Figure \ref{fig:model}.

\subsection{Results with SE Group Mixed Datasets}

\begin{figure*}[!ht]
\centering
\begin{tabular}{@{}cc@{}}
\subfloat[REFIT CtRNN]{\includegraphics[width=0.47\linewidth]{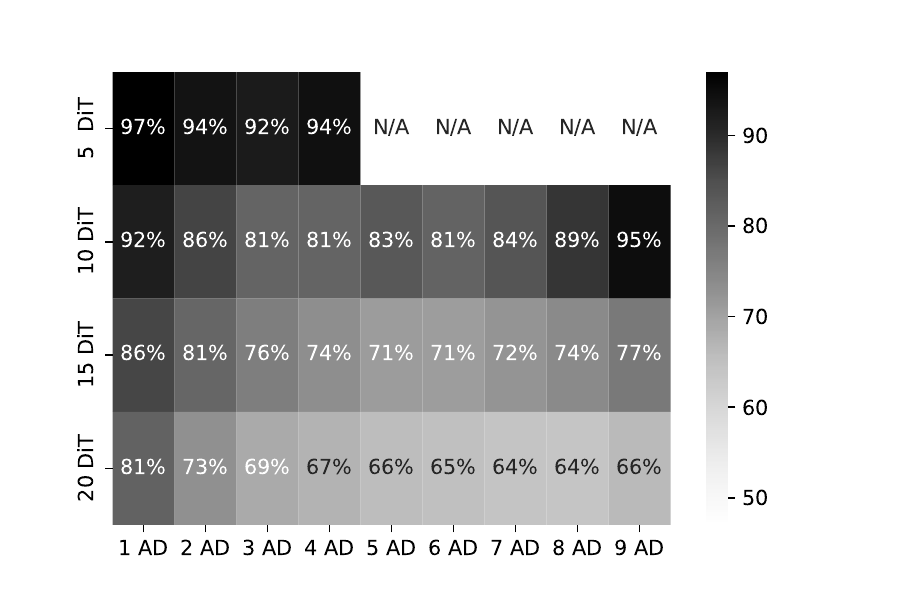}}\hspace{-4em} &
\subfloat[REFIT VGG11]{\includegraphics[width=0.47\linewidth]{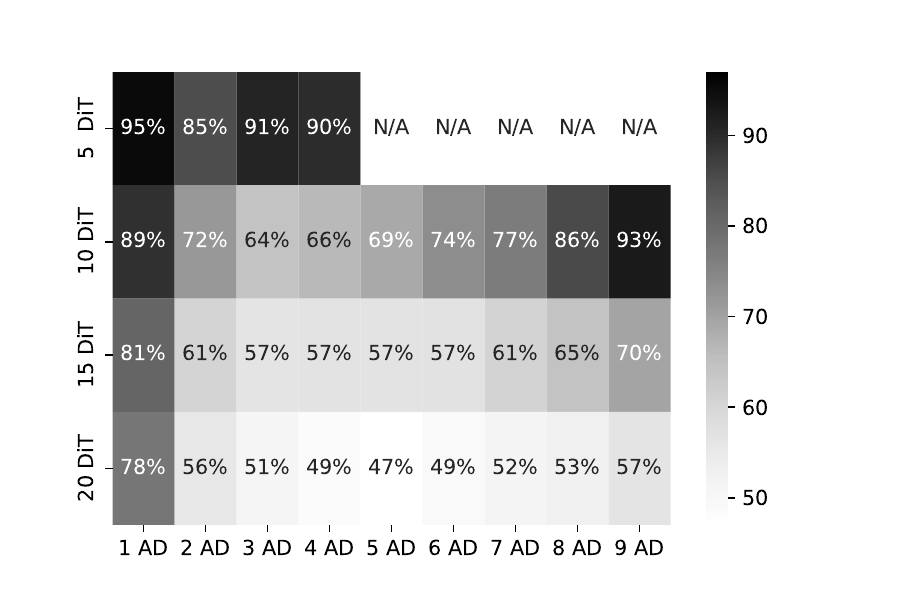}}\hspace{-4em} \\
\subfloat[REFIT RF]{\includegraphics[width=0.47\linewidth]{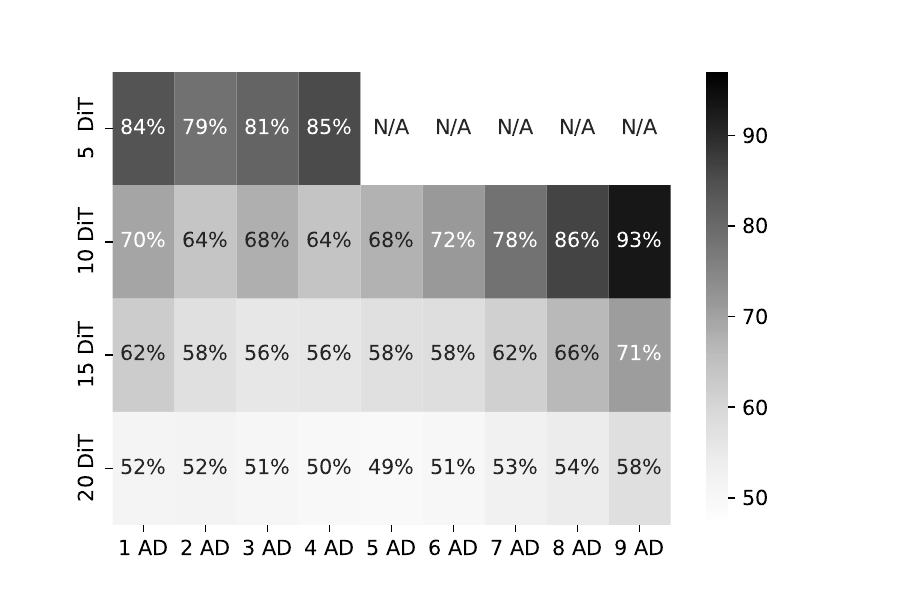}}\hspace{-4em} &
\subfloat[REFIT Random]{\includegraphics[width=0.47\linewidth]{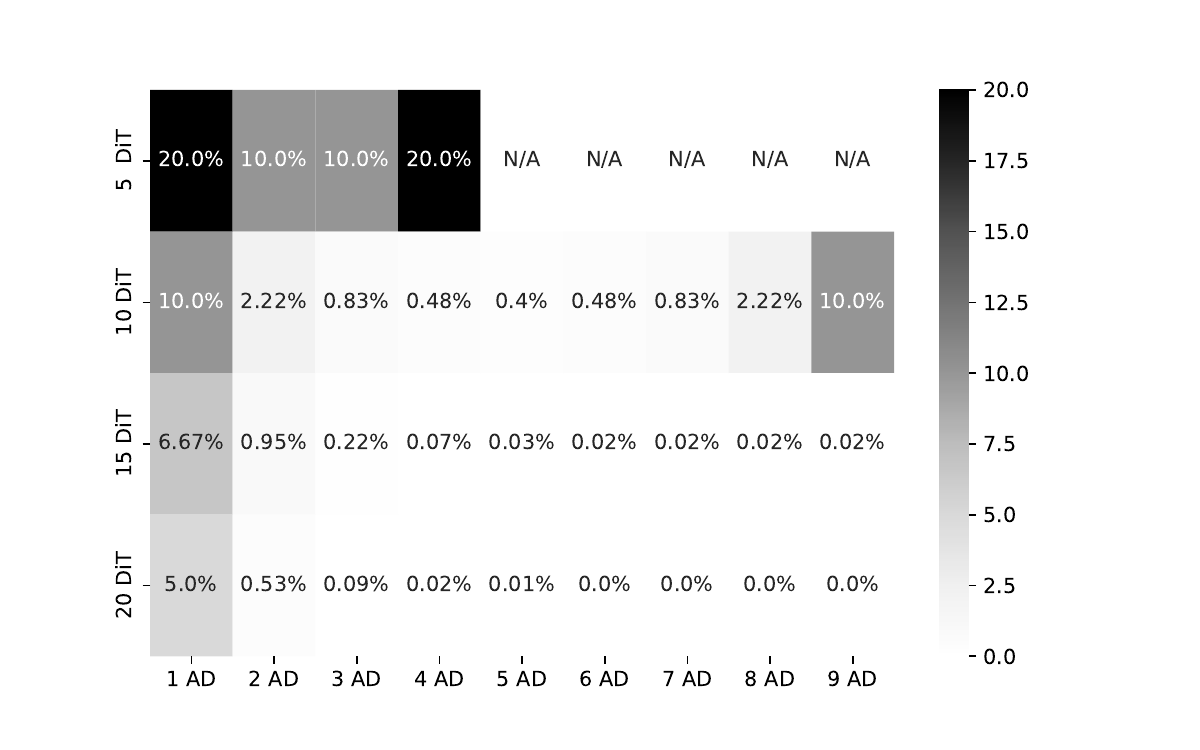}}\hspace{-4em} \\
\subfloat[UK-DALE CtRNN]{\includegraphics[width=0.5\linewidth]{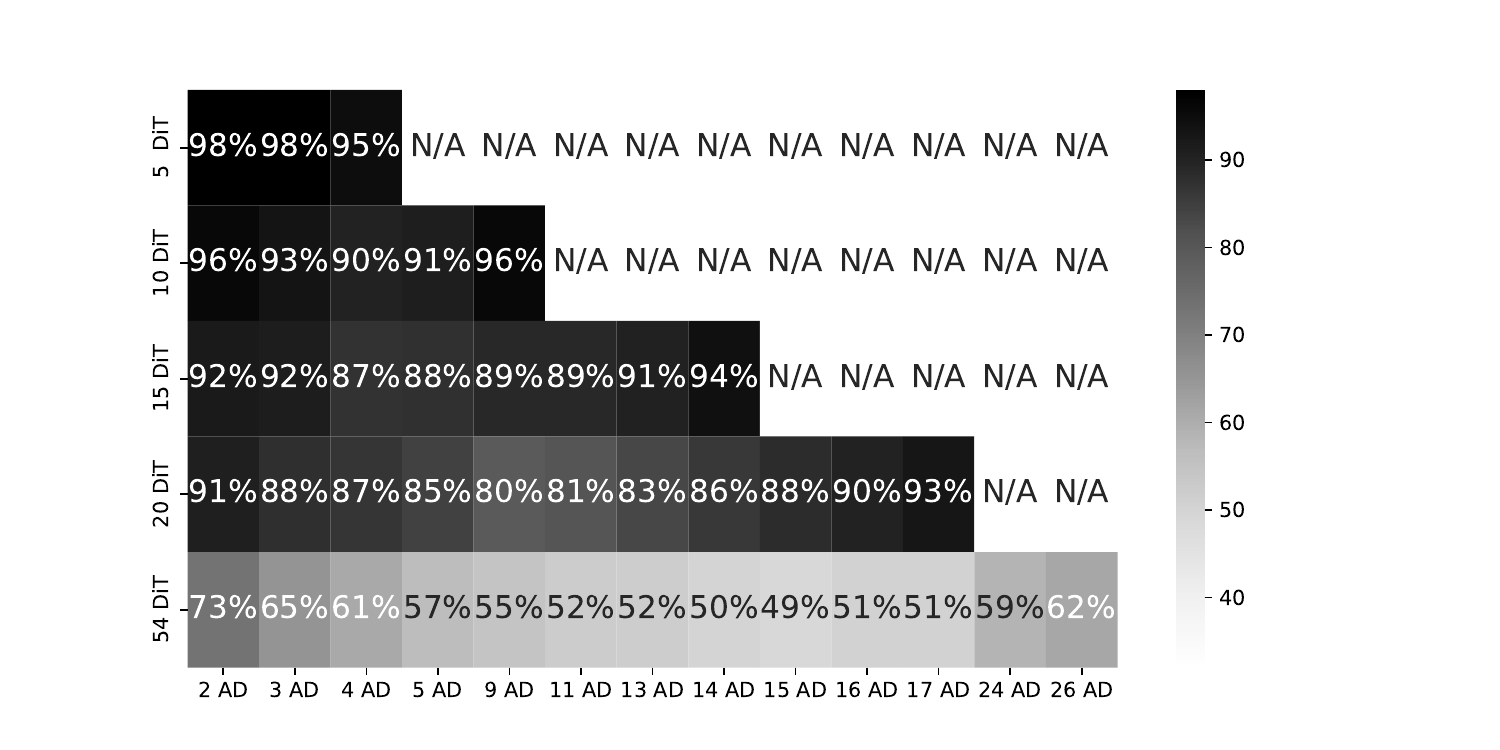}}\hspace{-4em} &
\subfloat[UK-DALE VGG11]{\includegraphics[width=0.5\linewidth]{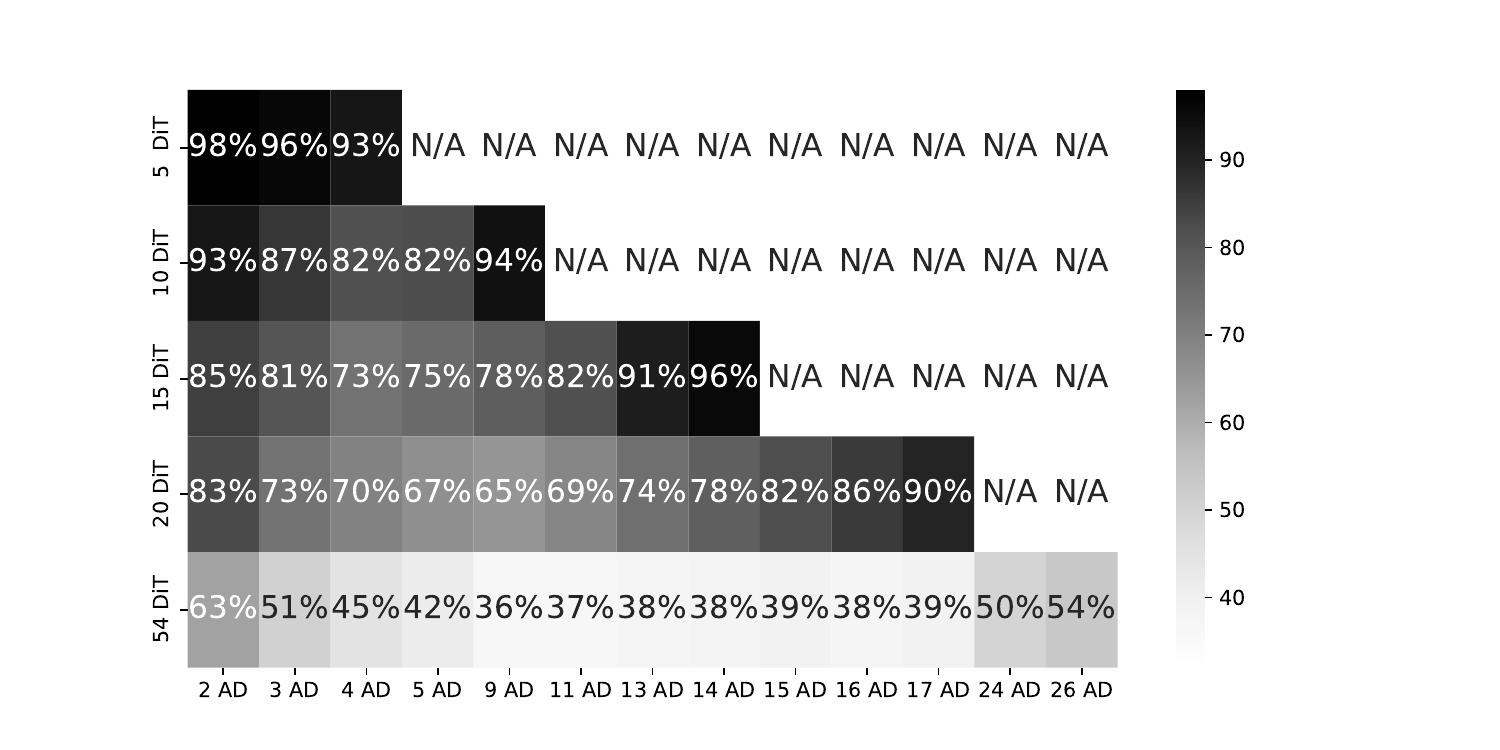}}\hspace{-4em} \\
\subfloat[UK-DALE RF]{\includegraphics[width=0.5\linewidth]{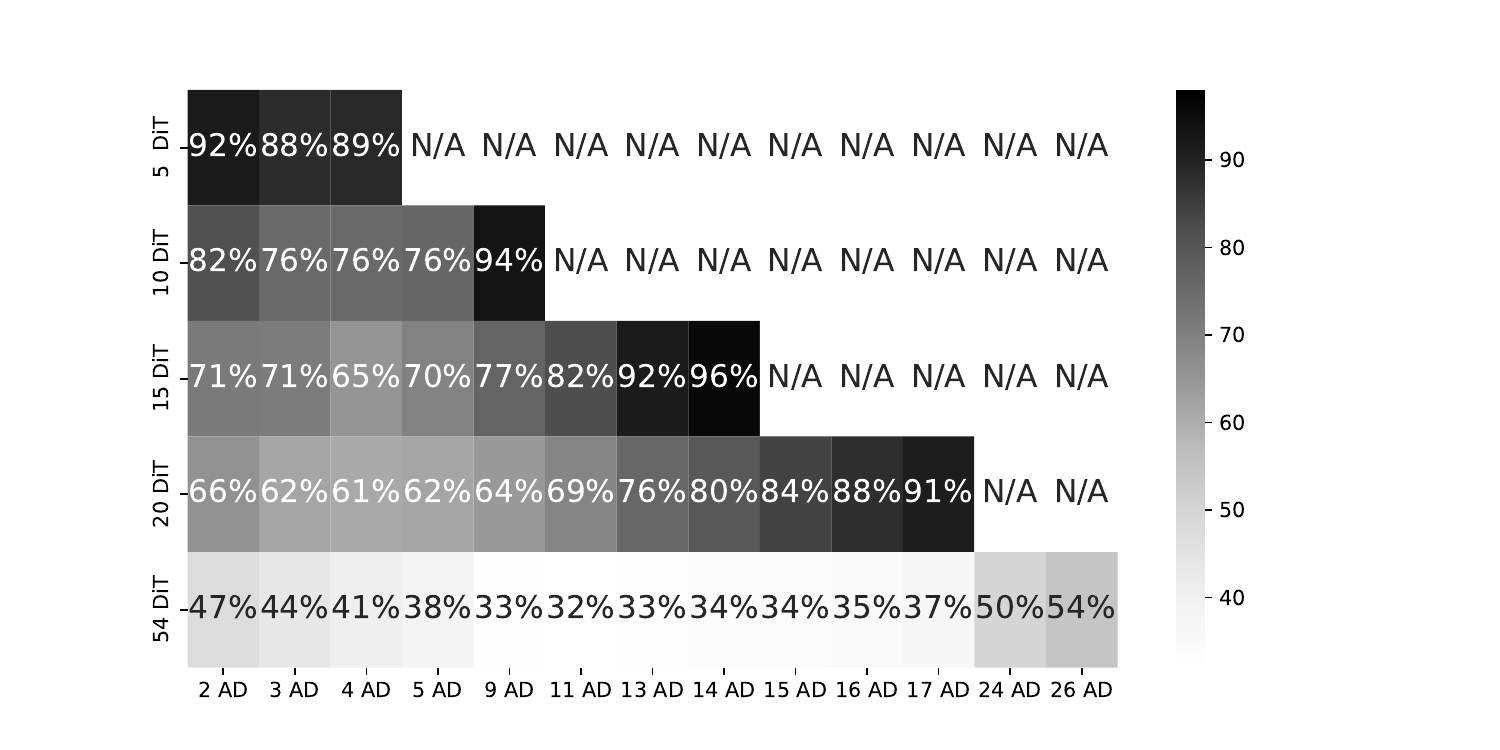}}\hspace{-4em} &
\subfloat[UK-DALE Random]{\includegraphics[width=0.6\linewidth]{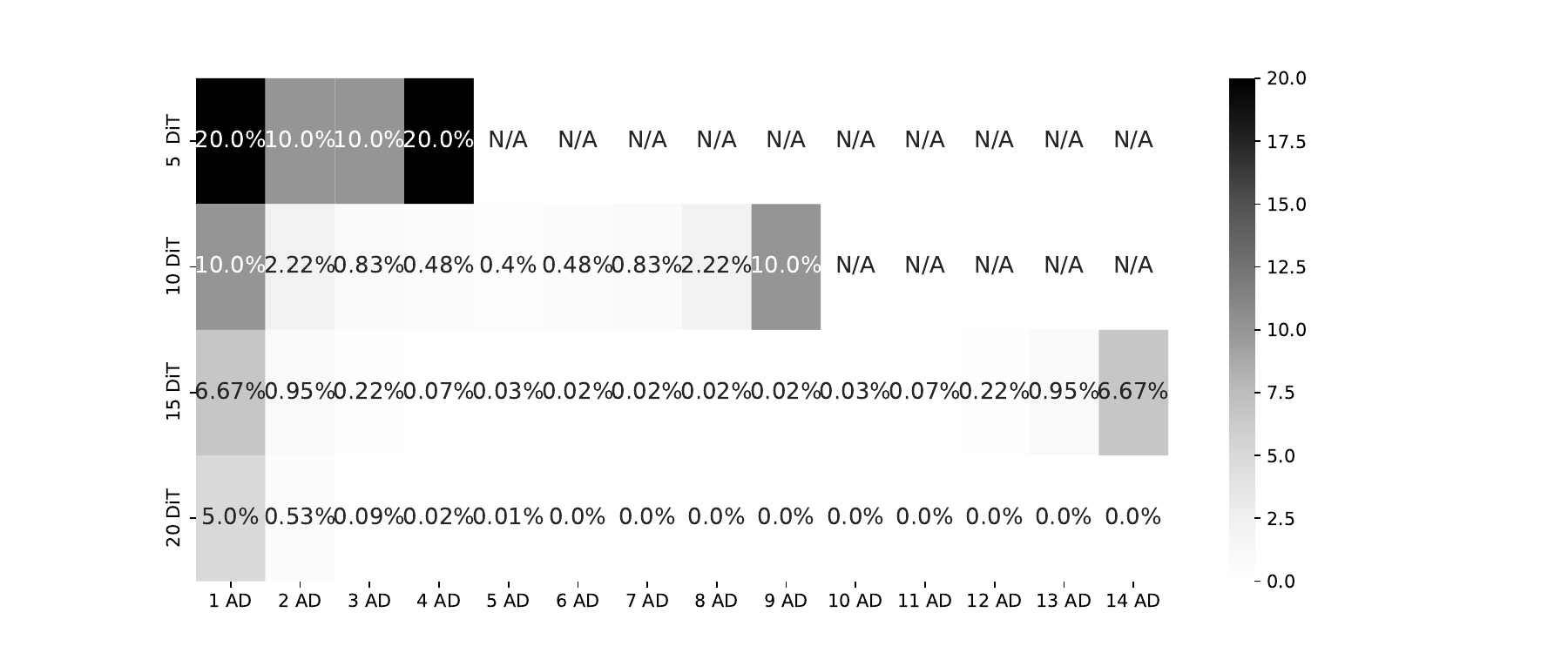}}\hspace{-4em} \\
\end{tabular}
\caption{Results from RF, VGG11 and CtRNN on the SE Group of mixed datasets.}
\label{fig:testA}
\end{figure*}

\begin{figure*}[htb]
    \centering
    \subfloat[REFIT]{
        \includegraphics[width=0.4\linewidth, clip, trim={0 0 0 0}]{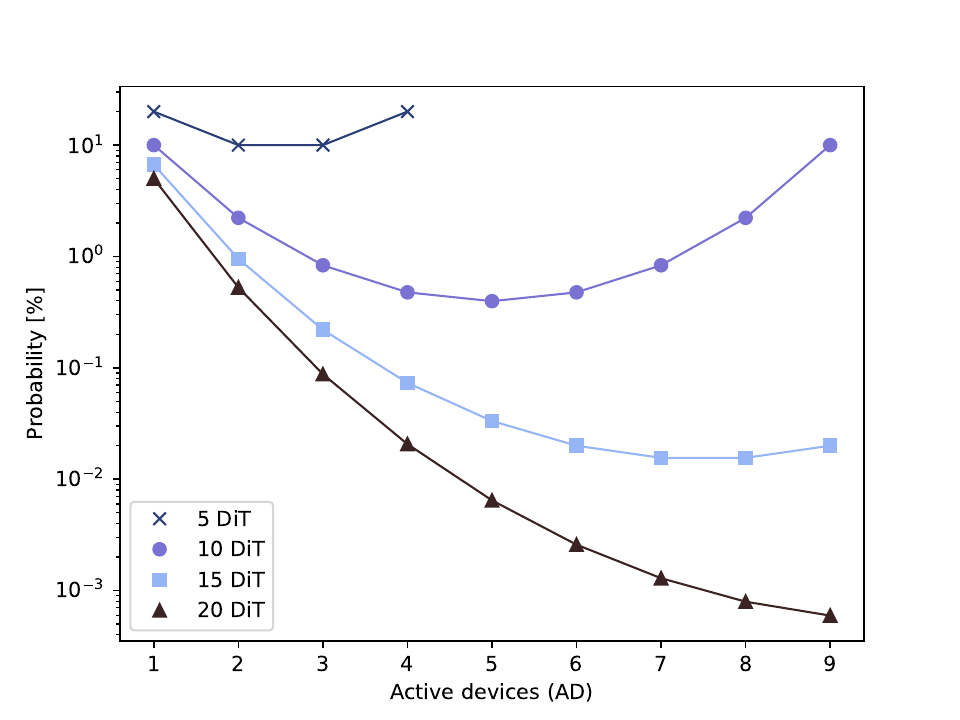}
    }
    \subfloat[UK-DALE]{
        \includegraphics[width=0.4\linewidth, clip, trim={0 0 0 0}]{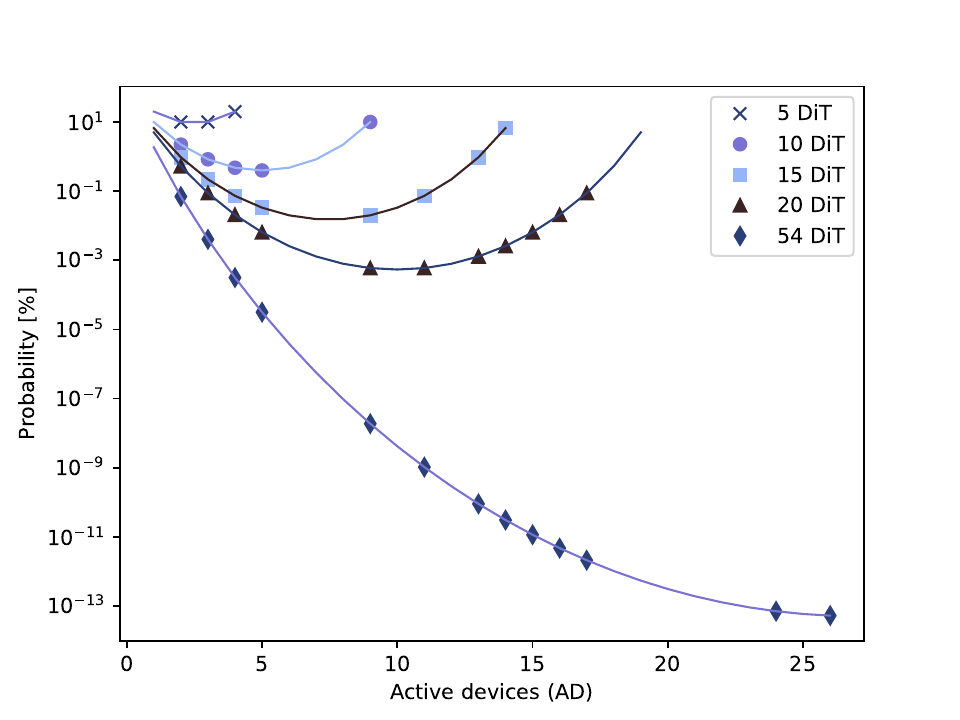}
    }
    \caption{Probability of accurately determining the status (ON/OFF) of devices through random chance. The impact of this chance is reflected in the results from RF, VGG11 and CtRNN on the SE Group of mixed datasets, which are partially numerically aligned with these probability lines.}
    \label{fig:probabilitycurve}
\end{figure*}

The results of comparing the performance of CtRNN, VGG11 and RF models on Sensitivity Evaluation Group mixed datasets are displayed in Figure~\ref{fig:testA}. Figures~\ref{fig:testA} a-c exhibit heatmaps presenting the evaluation results of the models on the REFIT dataset, whereas Figures~\ref{fig:testA}.e-g depict heatmaps displaying the evaluation results on the UK-DALE dataset. Figure~\ref{fig:testA}d and Figure~\ref{fig:testA}h, on the other hand, illustrate the probabilities of obtaining correct results as per random guess, calculated with Eq.~\ref{eqn:GroupAChance}.

\begin{equation}
\label{eqn:GroupAChance}
P_{GroupSE} = \frac{1}{\binom{DiT}{N_{AD}}}
\end{equation}

As it can be seen from  Figures~\ref{fig:testA}a-d, for the REFIT dataset, the accuracy of the model is affected both by the increase of the number of active devices (AD), as well as increase in the total number of devices (DiT) that can be active at the same time. It can be seen that the  average performance degradation of our model is 9.2 percentage points per each 5 DiT added on REFIT dataset and by 5.4 percentage points on UK-DALE dataset. However, as it can be seen for all three utilised classifiers and the theoretical calculation, the classification performance increases when the number of AD is approaching the number of DiT. All three classifier models significantly outperform the random classifier, with our approach CtRNN being the best out of the three. Looking at the the third row of heatmaps depicting results for 15 DiT in Figures~\ref{fig:testA}a-c, it can be seen that our approach achieves scores above 71\,\%, while models based on VGG11 and RF, achieve scores above 57\,\% and 56\,\%. All models significantly outperform the random chance as it reaches numbers as low as 0.02\,\%.

Similar observations can be also seen in Figures~\ref{fig:testA}e-h for the UK-DALE dataset. All three models significantly outperformed the random model, where again our approach achieved the highest score across all tested scenarios. Looking at the the third row of heatmaps depicting results for 15 DiT in Figures~\ref{fig:testA}e-g, it can be seen that our approach achieves slightly lower accuracy scores compared to the VGG11 and RF algorithms, when the number of AD approach the number of DiT, by up to 0.02. The reason for this is due to the fact that our approach is less prone to overfitting, compared to the other two approaches, which is supported by the fact that for up to 11 AD our approach significantly outperforms the other two approaches by up to 18 percentage points.

To summarize the difference between the results of different models we calculated the average improvement (I) across all mixed datasets in SE Group using Eq.~\ref{eqn:AvgImprovement}, our model outperforms the VGG11 model by 11.03 and 9.4 percentage points on the REFIT and UK-DALE derived datasets, respectively. Compared to the RF model, our model achieves even greater improvement with 14.15 and 13.88 percentage points on the REFIT and UK-DALE derived datasets, respectively.

\begin{equation}
\label{eqn:AvgImprovement}
\resizebox{0.8\linewidth}{!}{
$I = \frac{\sum_{n=1}^{N_{datasets}} (\overline{F1score_{w\_CtRNN_n}} - \overline{F1score_{w\_X_n}})}{N_{datasets}}$
}
\end{equation}

%[From the analysis of F4 you notice:]
From the analysis of heatmaps in Figure~\ref{fig:testA} we notice that once the number of AD surpasses 50\,\% of DiT in the mixed dataset, the chance of correct classification increases. This trend is clearly visible for both REFIT and UK-DALE in the lines with 5, 10 and 15 DiT and less so in the line with 20 DiT and for UK-DALE in the line with 54 DiT. To better understand the trend, we calculate the probability of correctly classifying devices by random guess in SE Group mixed datasets with Eq.~\ref{eqn:GroupAChance} and depict the results in Figure~\ref{fig:probabilitycurve}. The x-axis represents the number of active devices (AD) out of devices in total (DiT), while the y-axis represents the probability of guessing the results correctly. We display a curve for each number of DiT in the SE group, thus showcase the probability for all employed combinations of AD and DiT. Consider the curve for 5 DiT in Figure \ref{fig:probabilitycurve}a. It can be seen that when randomly picking 1 AD out of 5 DiT the likelihood of it being correct is 20\%. Next, the likelihood of correctly guessing 2AD of 5DiT is 10\%, 3 out of 5 is 10\% while 4 out of 4 is 20\%. The results in the figures show that, for a small number of AD or a number of AD that is comparable with the number of DiT, the random guess works the best, while for the cases when the number of AD is about 50\% of the number of DiT it yields the worst results. That is because probability is expressed with combinations as shown in Eq. \ref{eqn:GroupAChance}, since the order of predicted active devices doesn't matter. Looking at  Figure~\ref{fig:probabilitycurve} we notice similar decrease and increase in performance as previously seen in rows of the heatmaps in Figure~\ref{fig:testA}.

\begin{figure*}[htb]
    \centering
    \begin{tabular}{@{}cccc|cccc@{}}
    
    \multicolumn{4}{c|}{REFIT} 
    & \multicolumn{4}{c}{UK-DALE}  
    \\
    
    \cmidrule(lr){1-4}  
    \cmidrule(lr){5-8}  

    \subfloat[CtRNN]{\includegraphics[width=0.1\linewidth]{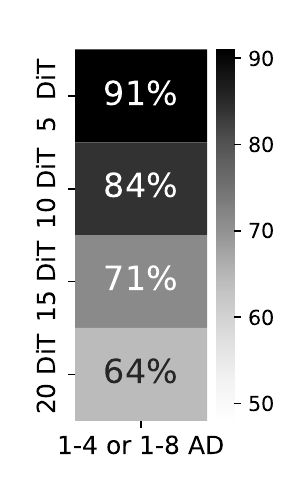}} &
    \subfloat[VGG11]{\includegraphics[width=0.1\linewidth]{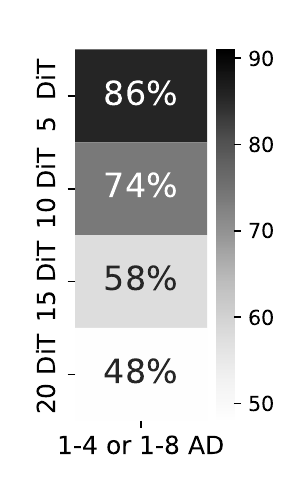}} &
    \subfloat[RF]{\includegraphics[width=0.1\linewidth]{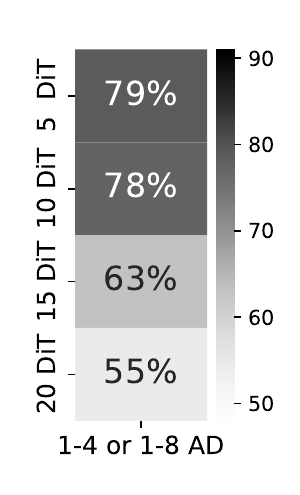}} &
    \subfloat[Random]{\includegraphics[width=0.1\linewidth]{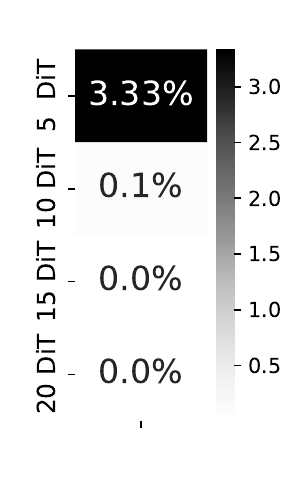}} &
    
    \subfloat[CtRNN]{\includegraphics[width=0.1\linewidth]{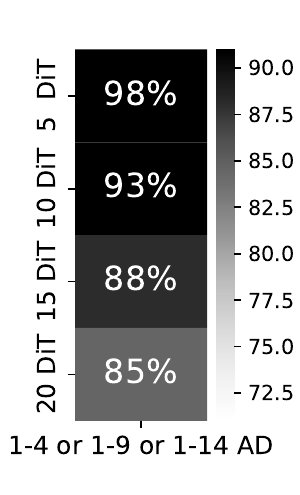}} &
    \subfloat[VGG11]{\includegraphics[width=0.1\linewidth]{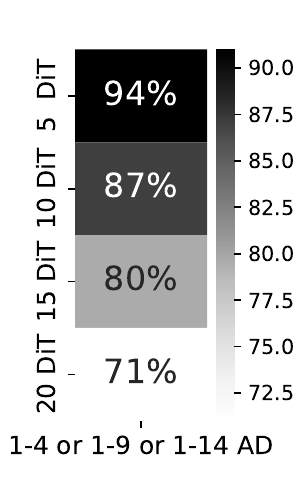}} &
    \subfloat[RF]{\includegraphics[width=0.1\linewidth]{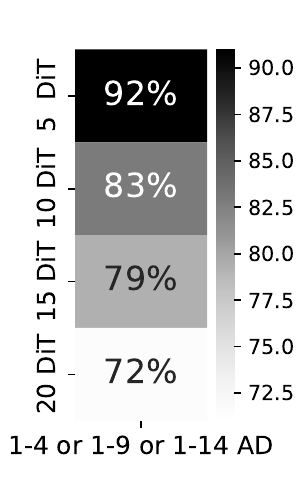}}&
    \subfloat[Random]{\includegraphics[width=0.1\linewidth]{figs/B/Combinatorics_B_REFIT_DALE.pdf}}
    
    \\
    
    \end{tabular}
    \caption{Results from RF, VGG11 and CtRNN on the RE Group of mixed datasets.}
    \label{fig:testB}
\end{figure*}

\subsection{Results with RE Group Mixed Datasets}
The results of comparing the performance of CtRNN, VGG11 and RF models on RE Group mixed datasets are displayed in Figure~\ref{fig:testB}, Figures~\ref{fig:testB}a-c showcase heatmaps presenting results from evaluation on REFIT, while Figures~\ref{fig:testB}e-g characterize heatmaps displaying results from evaluation on UK-DALE. Figures~\ref{fig:testB}d and h, contain heatmaps filled with probabilities of obtaining correct results randomly, calculated with Eq.~\ref{eqn:GroupBChance}.

We observe that with the increase in DiT the accuracy of classification decreases in all cases, which is consistent with our prior observations related to performance being connected with the proportion between ADs and DiTs. The average performance degradation of our model is 9 percentage points per each 5 DiT added on REFIT dataset and by 4.3 percentage points on UK-DALE dataset. Random probability of correct classification, calculated by Eq.~\ref{eqn:GroupBChance} is much lower compared to the accuracy of the models. For example, on REFIT dataset in the row with 15 DiT, our model achieves a score of 71\,\%, VGG11 achieves 58\,\% and RF achieves 63\,\%, whereas the random probability is rounded to 0\,\%. 

\begin{equation}
\label{eqn:GroupBChance}
P_{GroupRE} = \frac{1}{2^{DiT} - \binom{DiT}{0} - \binom{DiT}{N_{AD}+1} - . . . - \binom{DiT}{DiT}}
\end{equation}

We calculate the average improvement over the entire RE Group of mixed datasets using Eq.~\ref{eqn:AvgImprovement}. Our model reaches results that are 11.32 percentage points better than VGG11 and 9.22 percentage points better than RF on REFIT derived dataset, and 8.07 percentage points better than VGG11 and 9.46 percentage points better than RF on UK-DALE derived dataset, respectively.

\begin{table*}[htbp]
    \ra{1.3}
    \caption{Summary of related work for multi-label classification on NILM.}
    \label{tab:ComparisonWithRelatedWork}
    \centering
    \resizebox{\textwidth}{!}{
    \centering
    \begin{tabular}{l|c|c|cc|cc|c|c}
        \toprule
        \bfseries Work 
        & \bfseries Approach Type 
        & \bfseries Approach  
        & \multicolumn{4}{c|}{\bfseries Reported Evaluation} 
        & \bfseries Type
        & \bfseries Devices no.
        \\
        \cmidrule{4-7}
        & & &\bfseries Dataset & \bfseries avg. F1 & \bfseries Dataset & \bfseries avg. F1 &
        \\
        \midrule
        Tabatabaei~\textit{et al.}~\cite{Tabatabaei2017}
        & Classic ML 
        & MLkNN 
        & REDD 
        & 0.528
        & /
        & /
        & LF
        & 5
        \\
        Wu~\textit{et al.}~\cite{Wu2019}
        & Classic ML 
        & RF
        & BLUED 
        & 0.98
        & /
        & /
        & HF
        & 5
        \\
        Singh~\textit{et al.}~\cite{Singh2022}
        & Classic ML 
        & SRC
        & REDD
        & 0.70
        & Pecan Street 
        & 0.71
        & LF, LF
        & 4
        \\
        Tanoni~\textit{et al.}~\cite{Tanoni2022}
        & DL 
        & CRNN
        & REFIT
        & 0.83
        & UK-DALE 
        & 0.87
        & LF, LF
        & 5
        \\
        Langevin~\textit{et al.}~\cite{Langevin2022}
        & DL 
        & CRNN
        & REFIT
        & 0.78
        & UK-DALE 
        & 0.68
        & LF, LF
        & 5
        \\
        \midrule
        This work
        & DL 
        & CtRNN
        & REFIT
        & 0.91
        & UK-DALE 
        & 0.94
        & LF, LF
        & 5
        \\
    \bottomrule
    \end{tabular}
    }
\end{table*}

\subsection{End Performance Comparison}
\label{subsec:ComparisonOfResultsWithRelatedWorks}

Assuming a quick technology selection for an application requiring ON/OFF classification would need to be performed based only on the end performance, type of ML and number of devices, we summarize in Table~\ref{tab:ComparisonWithRelatedWork} the required information. To compile  Table~\ref{tab:ComparisonWithRelatedWork}, we take the related ON/OFF classification works summarized in Table~\ref{tab:RelatedWork} and analyzed in Section \ref{sec:related_work}, and we extract the best final results from the respective papers, except for Tanoni in the fourth row of the table where we present the results reported in this work. The exception for Tanoni is due to the fact that the original work \cite{Tanoni2022} employs weak supervision so the results are slightly worse then achieved with  supervised learning in all other works. For fairness to Tanoni, we re-rune their experiments in a fully supervised manner and report those results. 

The first column of the related works in  Table~\ref{tab:ComparisonWithRelatedWork} lists the considered ON/OFF classification works, the second lists the type of ML approach (classical or deep), the third provides the specific method, the fourth the dataset and result when training with it, the fifth lists the type of sampling of the energy data while the last lists the number of considered devices. From the results it can be seen that Wu et al. achieved the highest F1 score 98\% however this has been done on the HF dataset BLUED. When a signal is sampled with higher frequency, higher definition data is available therefore it is easier to recognize its shape compared to a signal that is sampled with less granularity. When compared to other similar models developed on LF data, our model reached scores around 92.5\%, surpassing others. TanoniCRNN ranked second on LF datasets, with scores around 0.85\% while Langevin et al. and Singh et al. had similar scores, approximately 0.7\%. The result reported by Tabatabei et al. ranked the lowest at 53\%.

\subsection{Limitations and Future Work}

The \textit{limitations}  of the study presented in this paper are twofold. First, we show that the approach is not robust to increasing number of devices which characterize  modern households. Unlike prior works, we quantify the drop in performance that we mostly attribute to the imbalanced nature of the available datasets in which some devices occur more frequently than others. Second, the empirical design of the  proposed architecture could be automated to find a superior architecture with respect to both performance and energy efficiency. 

While there are already a number of works on ON/OFF classification that improve on prior ones, including this study, we see three main lines of \textit{future work} as follows.

\paragraph*{Benchmark dataset} Typically in the machine learning communities they have benchmark datasets that are used in all model evaluations. A good direction for future work is to take the existing datasets that are suitable for ON/OFF classification and generate a harmonized set suitable for training. Besides the harmonized set, also balanced versions could be created,  using statistical oversampling or under-sampling methods such as ADASYN and AllKNN. Also a more general simulator than in \cite{klemenjak2020synthetic} could be developed to permit complementing measured data.

\paragraph*{Generic model development} 
Current ON/OFF classification models cannot just be downloaded and used in a real application. While there is recent work for transferring models across household they have limitations, especially reflecting in performance drop \cite{Bertalanič2022}. For text, the GPT breakthrough has shown that the architecture can be trained on large amounts of unlabelled data to capture sufficient knowledge and then further trained on labelled data to better structure that knowledge. The same could be done for ON/OFF classification where a model trained on large amounts of general time series data could be adapted for the domain at hand. The development of such a model could also significantly lower the adoption barrier of such technology.

\paragraph*{Role in smart energy management} 
According to \cite{ehrhardt2010advanced} households consume 12\% less energy if they receive specific feedback on the consumption of individual devices. Furthermore, increasingly automated energy management systems may rely on machine learning models to detect appliances and forecast usage. Quantification of the relationship between the performance of a ML model (i.e. F1-score, MSE), its energy consumption and the energy that its decision help saving is a worthy line of research. For instance, if on average the household consumption drops by 12\% \cite{ehrhardt2010advanced} and the model on average recognizes  only 94\%  of the devices correctly, as in this study, what would be the impact on the trust and behaviour of the user? Also, misclassifying an air conditioning device that consumes significant energy may have a higher cost than misclassifying a water heater.

\section{Conclusions}
\label{sec:conclusion}
In this paper, %we (i) propose a new DL architecture used in the CtRNN model, and (ii) address the problem of unrealistic performance evaluation of algorithms applied to NILM dissagregation and analyze the energy efficiency of different approaches of ON/OFF classification on NILM. 
%To minimize the downside of DL approaches, 
we propose a new DL architecture CtRNN used in our models, paying special attention to improve its energy efficiency during training and operation as well as its performance compared to state of the art and other similar models. %We developed CtRNN 
In developing the architecture, we used a typical VGG family architecture as a starting point and adapted it to time series data and reduced computational complexity by reducing the number of convolutional layers in some blocks, replacing one block with a single transposed convolution layer and adding the GRU layer. We benchmarked the proposed new model with other similar models, %Moreover, we evaluate performance of DL approaches and the best classical ML approach according to related work and conclude that DL approaches have a much higher performance potential.
%added conclusion:
showing that it is possible to develop new DL models for NILM ON/OFF classification that provide a major improvement in both performance and energy efficiency, which results in lesser energy consumption.

%To solve the second problem 
We also proposed a new methodology with two tests that more realistically assess the performance of NILM ON/OFF classification algorithms. They are using groups of multiple mixed datasets, derived from measurement datasets with specificities of the real-world use cases in mind. One group covers the numbers of active devices commonly used in the time window of the learning samples in separate datasets. The other group covers mixed number of devices from 1 to the average maximum number of devices used in the time window of the learning sample.
% added conclusion
Our findings demonstrate that the proposed methodology is necessary to obtain results, which reflect more realistic situations. The obtained results indicate that the commonly used testing methodology can lead to overly optimistic conclusions, underscoring the importance of employing a more rigorous evaluation framework.

%there is some conclusion in this one.
Moreover, as part of performance evaluation we also compared DL approaches and the best classical ML approach according to related work, and we concluded that DL approaches have a much higher performance potential compared to ML. In our experiments, we observed on average approximately 12 percentage point advantage of our model compared to the best classical ML approach.

\section*{Acknowledgments}
This work was funded in part by the Slovenian Research Agency under the grant P2-0016.
This project has received funding from the European Union’s Horizon Europe Framework Programme under grant agreement No 872525 (BD4OPEM).

\printglossary[type=\acronymtype]

\ifCLASSOPTIONcaptionsoff
  \newpage
\fi

\bibliographystyle{IEEEtran}
\bibliography{bibliography}

\end{document}